%% file: neurips_2026.tex
\documentclass{article}

\PassOptionsToPackage{numbers,compress}{natbib}
\usepackage[eandd, final]{neurips_2026}
\usepackage{wrapfig}
\usepackage{soul}
\usepackage{xcolor}
\usepackage{appendix}

\definecolor{bluelink}{rgb}{0.2, 0.81, 1}
\usepackage{threeparttable}
\usepackage[utf8]{inputenc} 
\usepackage[T1]{fontenc}    
\usepackage{hyperref}       
\usepackage{url}            
\usepackage{booktabs}       
\usepackage{amsfonts}       
\usepackage{nicefrac}       
\usepackage{microtype} 
\usepackage{tcolorbox}
\usepackage{multirow}
\usepackage{graphicx}
\usepackage{array}
\usepackage{makecell}
\usepackage{comment}
\usepackage{amsmath}
\usepackage{adjustbox}
\usepackage{graphicx}
\usepackage{wrapfig}
\usepackage{wrapfig}
\usepackage{float}
\usepackage{pgffor}
\usepackage{subcaption}
\usepackage{adjustbox}
\usepackage{longtable}
\usepackage{booktabs}
\usepackage{placeins}
\usepackage{algorithm}
\usepackage[noend]{algpseudocode}
\usepackage{caption}
\usepackage{amsmath}
\usepackage{amssymb}
\usepackage{ltablex}
\usepackage{booktabs}
\usepackage{tabularx}
\usepackage{makecell}
\usepackage{float} 
\keepXColumns
\usepackage{array} 
\usepackage[dvipsnames]{xcolor}
\usepackage{titlesec}
\usepackage{titletoc}
\usepackage[utf8]{inputenc} 
\usepackage[T1]{fontenc}    
\usepackage{hyperref}       
\usepackage{url}            
\usepackage{booktabs}       
\usepackage{amsfonts}       
\usepackage{nicefrac}       
\usepackage{microtype}      
\usepackage{xcolor}         

\newcommand{\colortextlink}{\color{bluelink}}
\usepackage{tikz}
\usepackage[normalem]{ulem}
\keepXColumns
\usepackage{array} 
\definecolor{darkgreen}{RGB}{17, 185, 17} 

\newcommand{\cmark}{\textcolor{darkgreen}{\checkmark}}
\newcommand{\xmark}{\textcolor{red}{\times}}

\newcommand{\graycircle}[1]{
  \tikz[baseline=(char.base)]{
    \node[shape=circle, draw=black, fill=gray, text=white, inner sep=1pt] (char) {\sffamily\bfseries\footnotesize #1};
  }
}

\title{\textit{Detect, Explain, Interpret:} \\ An End-to-End Benchmark for Time Series Anomaly Detection, Explainability and Interpretability.}

\author{
  Roberto Stanzione\\
  Inria, ENS, CNRS, PSL \\
  Paris, France\\
  \And
  Jules Barbe \\
  Scality \\
  Paris, France \\
  \And
  Magali Parrino \\
  Inria, ENS, CNRS, PSL, EDF \\
  Paris, France \\
  \And
  Jérémie Fourmann \\
  Scality \\
  Paris, France \\
  \And
  Paul Boniol\\
  Inria, ENS, CNRS, PSL \\
  Paris, France\\
}
\begin{document}

\maketitle
\begin{abstract}
 Time Series Anomaly Detection (TSAD) has received increasing attention, driven by the growing availability of complex time series data. This surge has led to the development of numerous detection methods, as well as a variety of benchmarks aimed at thoroughly evaluating their performance. However, most existing detectors remain largely agnostic to domain context, overlooking explainability and interpretability. One of the main reasons for this gap is that current benchmarks primarily focus on detection accuracy, and only few of them evaluate spatial explainability. Moreover, no benchmark currently provides sufficiently rich semantic annotations to support the generation of human-understandable interpretations of anomalies. To address these limitations, we introduce \textbf{SHAD} (\textbf{\underline{S}}cality \textbf{\underline{H}}igh-dimensional \textbf{\underline{A}}nomaly \textbf{\underline{D}}etection benchmark), a fully annotated benchmark composed of 215 multivariate, high-dimensional time series collected from real-world distributed cloud storage systems operated by Scality. The proposed dataset includes rich contextual information, covering three families of anomalies with varying degrees of severity. As further contribution, we provide a foundation for future work by evaluating baseline methods for \textit{Detection}, \textit{Explainability}, and \textit{Interpretability}, covering all stages of a TSAD pipeline.  For \textit{Detection}, we benchmark a wide range of existing anomaly detectors, testing their effectiveness on the proposed real-world dataset. Then, we consider \textit{explainability} by evaluating whether measuring the contribution of each dimension in the generated anomaly score can provide accurate anomaly attributions. Finally, for \textit{interpretability}, we investigate the effectiveness of frozen LLM baselines in localizing and interpreting anomalies, paving the way for future research towards interpretable and explainable TSAD.  
\end{abstract}

\section{Introduction}
The adoption of sensing solutions has enabled the collection of large amounts of  \textit{time series} \cite{palpanas2015data}, whose analysis is increasingly important across multiple domains \cite{bai2008forecasting,dadashova2021multivariate,ghassempour2014clustering}. In this context, \textit{time series anomaly detection} (TSAD) aims to identify patterns that deviate from normal behavior \cite{boniol2024dive}, and has attracted significant  attention over the years \cite{wenig2024series2graph++,antoni2019conditionIndicators,Gao2020ensemble}.
This has driven the introduction of several benchmarks~\cite{paparrizos2022tsb, ahmad2017unsupervised} which, however, focus on accuracy-oriented evaluation and overlook interpretability and explainability.
As a result, existing benchmarks lack rich semantic annotations, limiting the ability to assess whether models can meaningfully explain and interpret anomalies. The \textit{Exathlon} benchmark \cite{jacob2020exathlon} partially addresses this by providing root cause and effect intervals, but still uses coarse-grained anomaly types without explicit mappings to system components and lacks fine-grained, semantically rich per-dimension annotations. Thus, while the surge of LLMs is leading the definition of interpretable solutions \cite{gruver2023large, zhou2023one, zhuang2024see}, researchers highlight that the scarcity of time series data enriched with textual context forces them to rely on synthetic data \cite{xie2024chatts}, hindering a proper, real-world evaluation of the proposed approaches. As mentioned in \cite{xie2024chatts}:

\begin{tcolorbox}[colback=gray!5!white,colframe=gray!80!white,boxsep=1pt,top=1pt,bottom=1pt]
\textit{"However, research [...] for time series understanding and reasoning remains limited, primarily due to the scarcity
of high-quality datasets that align time series with textual information"}.
\end{tcolorbox}

\begin{figure}[tb]
\centering
\includegraphics[width=0.9\textwidth]{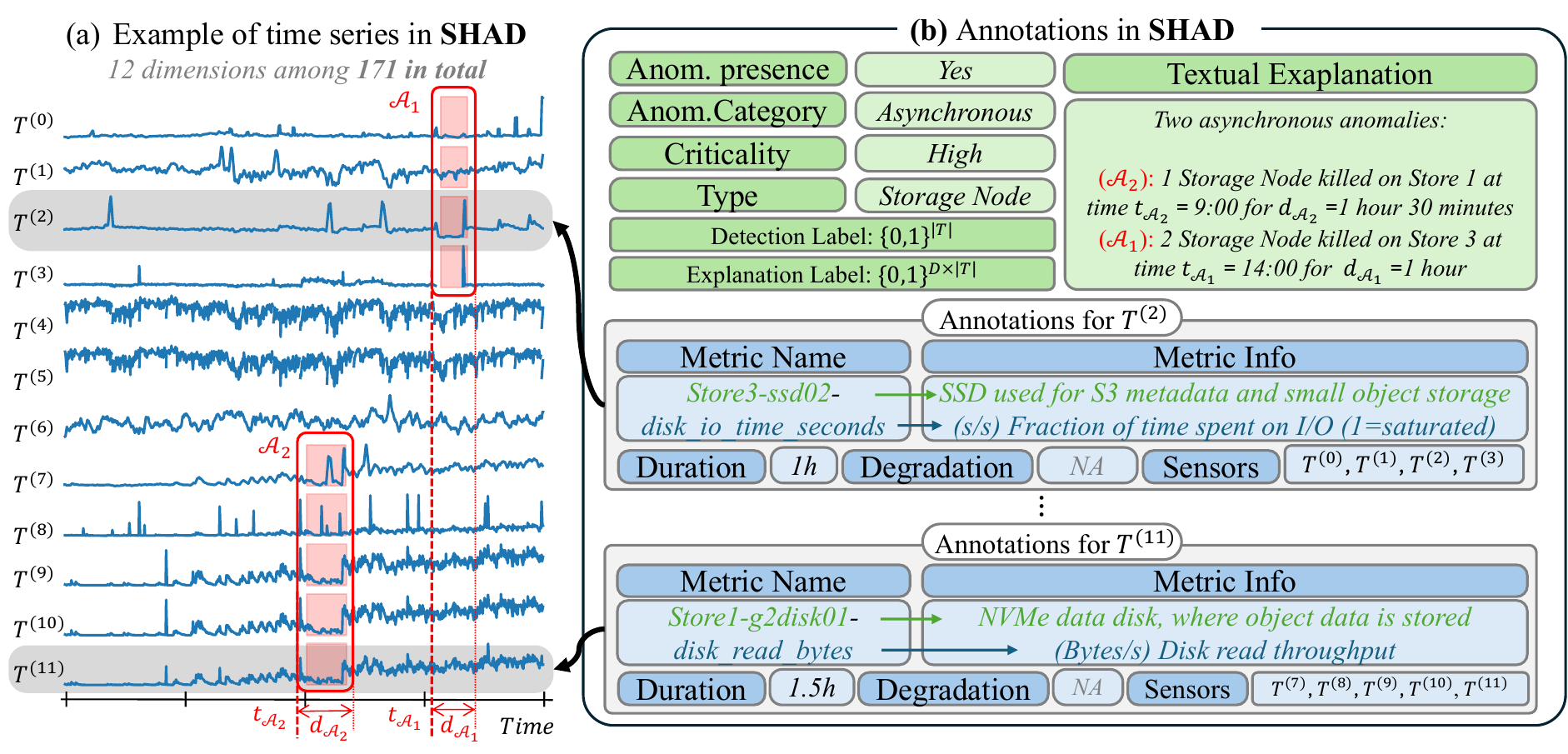}
\vspace{-0.2cm}
\caption{Example of SHAD's contextual knowledge (Asynchronous Storage Node Failure).}
\vspace{-0.4cm}
\label{fig:metadata_example}
\end{figure}

To address this gap, we introduce \textbf{SHAD} (\textbf{\underline{S}}cality \textbf{\underline{H}}igh-dimensional \textbf{\underline{A}}nomaly \textbf{\underline{D}}etection benchmark), a benchmark built from 
Scality RING distributed storage platform. SHAD includes both normal and anomalous series, with curated anomalies affecting different architectural layers. 
The dataset comprises $215$ high-dimensional time series with per-dimension labels and detailed documentation (see Figure \ref{fig:metadata_example}) describing anomaly type, affected components, and severity. Our public  repository\footnote{https://github.com/scality/shad/blob/main/README.md} also contains a detailed documentation of each of the 171 dimensions characterizing the series, enabling explicit mapping of each sensor to system components and roles. This level of semantic richness enables, for the first time, the evaluation of a complete TSAD pipeline, i.e., \textit{detection, explanation, interpretation}.
To summarize, the contributions of this paper are as follows: \textbf{(i)} we propose a novel benchmark composed of 215 high-dimensionality time series, with both nominal and anomalous behaviour, which includes rich per-dimension annotations and labels; \textbf{(ii)} we evaluate the anomaly detection performance of $18$ anomaly detectors on the proposed benchmark; \textbf{(iii)} we analyze spatial explainability by measuring the contribution of each component in the detected anomaly, and \textbf{(iv)} we evaluate the capability of zero-shot LLM baselines in localizing and explaining the anomalies.

\section{Background and Related Work} \label{sec:related}
In this section, we provide an overview of TSAD approaches by considering \textit{detection, explainability, and interpretability}. Then, we summarize existing benchmarks, as well as their limitations.

\paragraph{Notations.}
A \textit{multivariate time series} $ T \in \mathbb{R}^{(D, n)} $ is defined as a sequence of $ n $ observations across $ D $ dimensions, written $T = [T^{(0)}, T^{(1)}, \dots, T^{(D-1)}]$,
where $ T^{(i)} = [T^{(i)}_0, T^{(i)}_1, \dots, T^{(i)}_{n-1}] $ represents the $ i $-th dimension of $ T $. 
For instance, Figure \ref{fig:metadata_example} depicts a multivariate time series $T$ composed of $12$ dimensions.
Given a time series \(T\) and an anomaly detector \(D\), the detector outputs an anomaly score sequence
\(
S_T = [s_{0}, s_{1}, \ldots, s_{n-1}].
\)

\begin{figure}[tb]
\centering
\includegraphics[width=\textwidth]{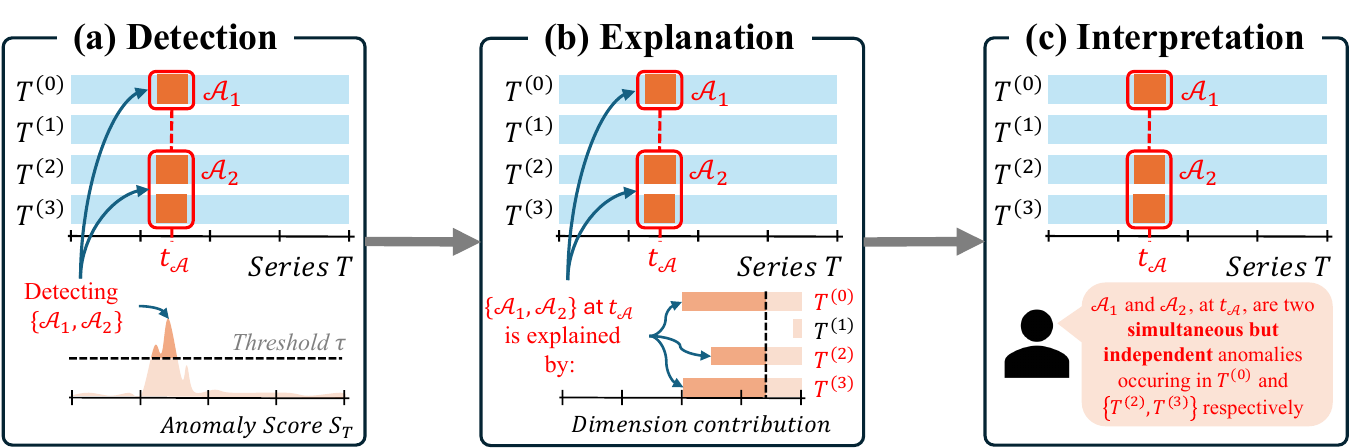}
\caption{Time series anomaly (a) \textbf{Detection}, (b) \textbf{Explanation}, and (c) \textbf{Interpretation}.}
\vspace{-0.4cm}
\label{fig:scope}
\end{figure}

\paragraph{Time Series Anomaly Detection.}
As illustrated in Figure~\ref{fig:scope}(a), the objective of anomaly detection is to temporally locate anomalies, independently of which dimension caused it.
Anomaly detectors can be grouped into the following categories.
\textit{Distance-based} methods identify anomalous subsequences by comparing them against their nearest neighbors in the series or in a reference set~\cite{keogh2007compression,liu2009efficient,yeh2016matrix,linardi2020matrix}.
\textit{Density-based} methods instead detect anomalous behaviors by estimating how isolated observations or subsequences are in a representation space~\cite{antoni2019conditionIndicators,boniol2020series2graph,liu2008isolation,Gao2020ensemble}. 
Finally, \textit{prediction-based} methods (\textit{forecasting}~\cite{Wu_2022} or \textit{reconstruction-based}~\cite{Omnianomaly}) estimate expected temporal behavior and use prediction error as the anomaly signal. Despite the large availability of anomaly detectors, most of them remain agnostic, therefore, achieving interpretability on a given domain is not straightforward. 

\paragraph{Interpretable and Explainable Anomaly Detection.}
The concerns about the black-box nature of deep-learning based approaches have led to efforts to make model behaviours more transparent and human-understandable~\cite{heuillet2021explainability}. Such studies usually refer to the notions of \textit{interpretability} and \textit{explainability}. However, these two terms are often used interchangeably, without a clear distinction. Thus, we now define these concepts as they are used throughout this paper,  by providing informal definitions (illustrated in Figure~\ref{fig:scope}(a,b)), as they inherently relate to human understanding.

\begin{tcolorbox}[colback=gray!5!white,colframe=gray!80!white,boxsep=1pt,top=1pt,bottom=1pt]
\textit{\textbf{Explainability:} We define explainability as the ability of a model to provide spatially grounded attributions that highlight which components of the input contribute to the detected anomaly.}
\end{tcolorbox}

Explainable time series anomaly detection methods are broadly divided into \textit{intrinsically explainable models} and \textit{model-agnostic} approaches~\cite{jacob2020exathlon}. The former build human-understandable representations (e.g., \textit{decision trees}) directly from data~\cite{molnar2020interpretable}, while the latter decouple explanations from the underlying model~\cite{ribeiro2016should}. In TSAD, forecasting and reconstruction methods~\cite{Wu_2022,Omnianomaly} are inherently explainable. Among model-agnostic techniques, some rely on input perturbations to observe changes in model outputs~\cite{lakkaraju2016interpretable}. LIME~\cite{ribeiro2016should} explains individual predictions via a local surrogate model learned through weighted linear regression, while Anchors~\cite{ribeiro2018anchors} extends this idea using logical rules. Other methods assign feature importance scores~\cite{bertossi2020causality, sundararajan2017axiomatic}. Among these,   SHAP~\cite{lundberg2017unified} is based on Shapley values~\cite{winter2002shapley}, and has been extended to time series~\cite{nayebi2023windowshap}. 
In~\cite{lee2023duogat} the authors propose DuoGAT, a TSAD approach based on a directed and weighted graph structure, and provide an explainability approach by hierarchically tracking critical time points to provide time-oriented reasoning. 
Finally, global explanation approaches approximate  models with explainable ones~\cite{frosst2017distilling, dong2024interpretability} or extract decision rules~\cite{kopp2020anomaly, lakkaraju2016interpretable, chen2025novel}.

\begin{tcolorbox}[colback=gray!5!white,colframe=gray!80!white,boxsep=1pt,top=1pt,bottom=1pt]
\textit{\textbf{Interpretability:} We define interpretability as the ability to produce a  human-understandable interpretation of the nature and context of anomalies.}
\end{tcolorbox}
Interpretable approaches remain relatively limited.  MacroBase~\cite{bailis2017macrobase} helps users focus on relevant portions of high-volume data streams, providing interpretable insights based on key attributes recurrence.
In the biomedical domain, Boubekki et al.~\cite{boubekki2026explaining} propose a method for interpreting models for ECG analysis through time-localized clustering of internal representations. Their approach associates temporally localized patterns with clinically meaningful conditions, supporting the interpretation of what the detected abnormal patterns represent.
Advances in large language models (LLMs) offer a promising direction, enabling context-rich interpretations by leveraging model outputs and auxiliary information~\cite{jin2024position}. While many works use LLMs for anomaly detection itself~\cite{jin2023time,wang2024news,rasul2023lag}, others explore their role in human-understandable data analysis~\cite{xing2021deepsqa, oh2023ecg}. LLMs have also been applied to integrate time series with external textual context, improving prediction and decision-making~\cite{kurisinkel2024text2timeseries, lai2025llmlight}, while agent-based and tool-augmented frameworks enhance reasoning and root cause analysis~\cite{wang2024rcagent, zhou2023d}. Among existing proposals~\cite{gruver2023large, zhou2023one}, only a few support multimodal data~\cite{zhuang2024see, xie2024chatts}. Among these, ChatTS~\cite{xie2024chatts} is a recent multimodal LLM capable of reasoning over multivariate time series and enables natural language interaction to interpret patterns, relationships, and anomalies.

\paragraph{Existing Benchmarks.} 
Research in TSAD has introduced numerous datasets (see Appendix~\ref{app:datasets}), reviewed in several surveys~\cite{paparrizos2022tsb, schmidl2022anomaly,sylligardos2023choose,liu2024elephant}, all highlighting the lack of a one-size-fits-all detector. This motivates the development of benchmarks (Table~\ref{tab:benchmark_comparison}) with heterogeneous datasets. 
\textbf{NAB}~\cite{lavin2015evaluating} focuses on real-time anomaly detection in streaming data, while \textbf{TODS}~\cite{lai2021revisiting} considers five anomaly types. \textbf{TimeEval}~\cite{schmidl2022anomaly} collects real and synthetic, univariate and multivariate datasets with point and sequence anomalies. \textbf{TimeSeAD}~\cite{wagner2023timesead} introduces a recall-consistent metric accounting for temporal correlations and benchmarks 28 deep learning methods, whereas \textbf{Experimental-TSAD}~\cite{zhang2023experimental} compares 17 SOTA
\begin{wraptable}[14]{r}{7.9cm}
  \scriptsize
  \centering
  \vspace{-0.3cm}
  \caption{Comparison of TSAD  benchmarks}
    \begin{tabular}{l|ccc|ccc}
    \toprule
    \multirow{2}*{\textbf{Benchmark}} & \multicolumn{3}{c}{\textbf{Dataset}} & \multicolumn{3}{c}{\textbf{Evaluation}} \\
    & Multi. & Ann. & Desc. & AD & Expl. & Interp. \\
    \midrule
    NAB~\cite{lavin2015evaluating} & $\xmark$ & $\xmark$ & $\xmark$ & $\cmark$ & $\xmark$ & $\xmark$  \\
    Exathlon~\cite{jacob2020exathlon} & $\cmark$ & $\xmark$ &  $\xmark$ & $\cmark$ & $\cmark$ & $\xmark$ \\
    TODS~\cite{lai2021revisiting} & $\cmark$  & $\xmark$ &  $\xmark$ & $\cmark$ & $\xmark$ & $\xmark$ \\
    TimeEval~\cite{schmidl2022anomaly} & $\cmark$  & $\xmark$&  $\xmark$ & $\cmark$ & $\xmark$ & $\xmark$ \\
    TSB-UAD~\cite{paparrizos2022tsb} & $\xmark$ & $\xmark$ & $\xmark$ & $\cmark$ & $\xmark$ & $\xmark$\\
    TimeSeAD~\cite{wagner2023timesead} & $\cmark$  & $\xmark$ &  $\xmark$ & $\cmark$ & $\xmark$ & $\xmark$  \\
    Exp-TSAD~\cite{zhang2023experimental} & $\cmark$ & $\xmark$ &  $\xmark$ & $\cmark$ & $\xmark$ & $\xmark$ \\
    TSB-AD~\cite{liu2024elephant} & $\cmark$  & $\xmark$ &  $\xmark$ & $\cmark$ & $\xmark$ & $\xmark$ \\
    TAB~\cite{qiu2025tab} & $\cmark$ &$\xmark$ &  $\xmark$ & $\cmark$ & $\xmark$ & $\xmark$  \\
    \midrule
    \textbf{SHAD (ours)} & \textbf{$\cmark$} & \textbf{$\cmark$} & \textbf{$\cmark$} &\textbf{$\cmark$} & \textbf{$\cmark$} & \textbf{$\cmark$} \\
    \bottomrule
    \end{tabular}
    \begin{tablenotes}
    \scriptsize
    \centering
      \item \textbf{Ann.:}  per-dimension  labels $\:\:$ \textbf{Desc.:} per-dimension description \\ \textbf{AD:} evaluation of anomaly detection performance
    \end{tablenotes}
  \label{tab:benchmark_comparison}
\end{wraptable}methods using both point and range-based metrics.
\textbf{TSB-UAD}~\cite{paparrizos2022tsb} provides a comprehensive benchmark for univariate TSAD with real and synthetic data. \textbf{Exathlon}~\cite{jacob2020exathlon} targets explainable anomaly detection in high-dimensional time series. To improve benchmark representativeness, \textbf{TSB-AD}~\cite{liu2024elephant} introduces a large-scale curated dataset and incorporates foundation model–based methods. Finally, \textbf{TAB}~\cite{qiu2025tab} offers a recent large-scale comparison across 44 anomaly detection approaches.

\paragraph{Limitations of Existing Benchmarks.}
Despite the available benchmarks, there are several limitations that need to be addressed:

\noindent
\textbf{L1} (\underline{\textit{Lack of end-to-end benchmarks}}):  
existing benchmarks mainly focus on anomaly \textit{detection} itself, thoroughly evaluating the detectors' accuracy but overlooking \textit{explainability} and \textit{interpretability}, which are critical parts of a TSAD pipeline. 

\noindent
\textbf{L2} (\underline{\textit{Lack of fine-grained anomaly labels}}):  
existing benchmarks only provide binary labels indicating whether an anomaly is present at a given time step, without specifying which dimension or sensor is affected, nor characterizing the nature of the anomaly itself.

\noindent
\textbf{L3} (\underline{\textit{Lack of contextual knowledge}}): existing benchmarks do not include rich semantic annotations and contextual information, hindering the generation of human-readable anomaly interpretations.  

\paragraph{Objectives and Research Questions.}
In response to these gaps 
we propose a real-world, high-dimensional curated dataset with rich anomaly descriptions and and  fine-grained documentation. We also include an experimental evaluation designed to answer the following research questions:

\noindent
\textbf{RQ1} (\underline{\textit{Detection}}): How do current state-of-the-art time series anomaly detection methods perform on a real-world, high-dimensional benchmark, and how do they compare in terms of effectiveness?

\noindent
\textbf{RQ2} (\underline{\textit{Explanation}}):  Can TSAD methods achieve explainability through dimension-level attribution?

\noindent
\textbf{RQ3} (\underline{\textit{Interpretation}}): Can frozen LLM models provide accurate anomaly interpretations? 

\section{SHAD: \textit{Our Proposed Benchmark}}\label{sec:Benchmark}
This section describes the system architecture underlying SHAD, the experiment design and anomaly simulation methodology, as well as specific dataset characteristics and labelling schema.

\textbf{System Overview.} To generate the datasets in SHAD, we deployed an experimental framework on \textit{{\colortextlink \href{https://www.scality.com/ring/}{Scality Ring}}}, which is a distributed object storage system in which data placement is governed by a \textit{Distributed Hash Table} (DHT) implemented via the \textit{Chord} protocol \cite{stoica2003chord}. In this architecture, software-defined storage nodes are responsible for specific segments of the keyspace and handle read and write operations for objects whose hashes fall within their assigned ranges. 
All clusters used to generate the dataset share a common topology (depicted in Figure \ref{app:topology} in the Appendix). Each deployment consists of a supervisor server responsible for management, monitoring, and orchestration of the experimental framework, along with three storage servers. Each storage server is equipped with four disks dedicated to data, two disks for metadata, and one disk hosting the operating system. On each storage server, six software storage nodes participate in the DHT, alongside a single S3 connector that manages client-facing object traffic and dispatches requests to the appropriate storage nodes.
Telemetry generated by all components is collected and organized into 171 metrics. 
A comprehensive description of all metrics is provided in Appendix~\ref{app:metrics}.

\begin{figure}[tb]
\centering
\includegraphics[width=0.95\textwidth]{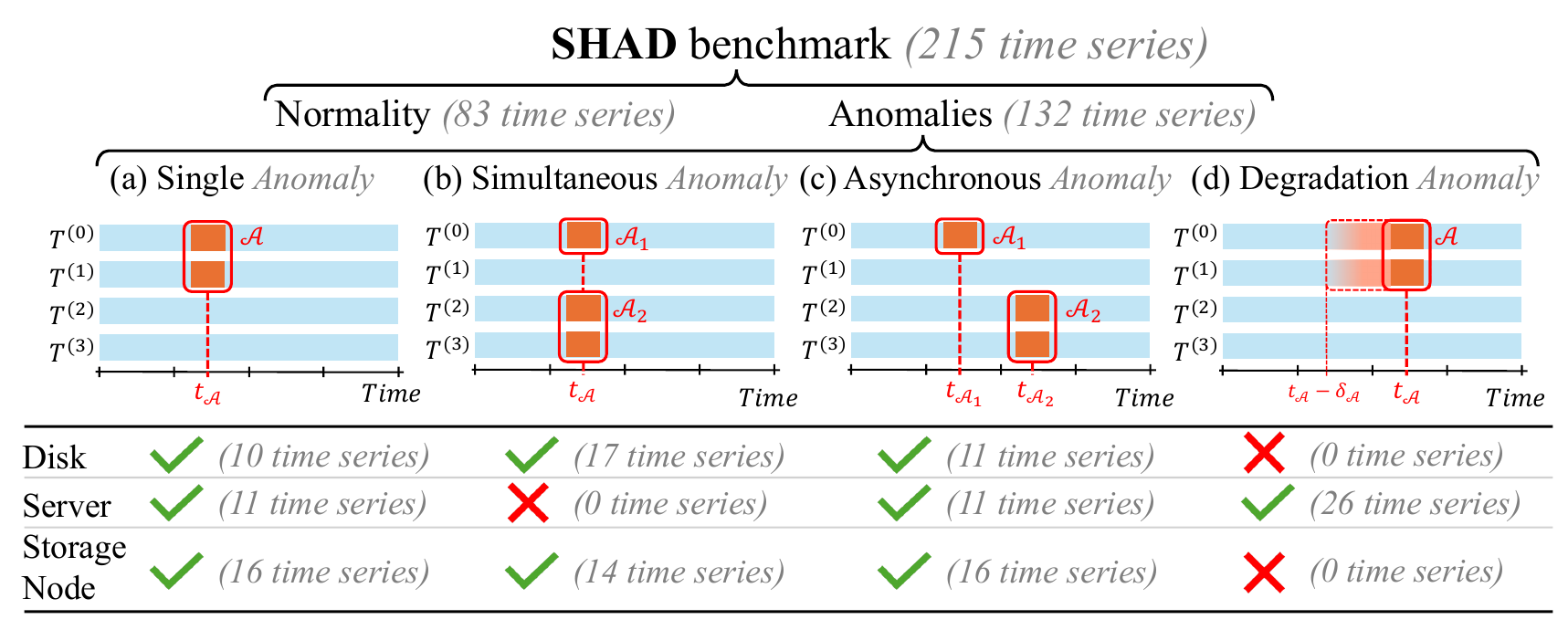}
\vspace{-0.3cm}
\caption{Overview of our proposed benchmark \textbf{SHAD}.}
\vspace{-0.6cm}
\label{fig:shad_overview}
\end{figure}

\noindent
\textbf{Experiment Design.} Our experiment framework runs one specific baseline workload consistent across all experiments. The workload runs uninterrupted for 18 hours with a one-minute telemetry ingestion interval, yielding 1000--1050 data points per experiment. The workload generates GET-heavy (i.e., dominated by object retrieval requests) mixed S3 traffic via the \textit{warp} benchmarking tool, using 16 concurrent workers, 15k initial objects at experiment start, 100KiB max randomized object size, and an S3 operation distribution of approximately 95\% \texttt{GET}, 4\% \texttt{STAT}, 0.8\% \texttt{PUT}, and 0.4\% \texttt{DELETE} by weight.
All experiments are conducted on top of this baseline workload, which simulates a representative workload of a typical storage cluster serving read-heavy object retrieval.

The experimental framework enables the controlled simulation of anomalies that affect otherwise nominal system traffic. All anomalous events are defined by their trigger time, their total duration, their targeted components, and extra failure-specific parameters. SHAD includes 3 general anomalies: \textit{server failures}, \textit{disk failures},  and \textit{storage node failures} (see Figure~\ref{fig:shad_overview}).

\textbf{\textit{(i) Disk failures}} are triggered by \textit{instantly} severing a server's connection with the targeted data disk(s).
\textit{Single} disk failures target one disk on a specific server. \textit{Simultaneous} failures target multiple disks at once,
while \textit{Asynchronous} failures target multiple disks at different timestamps. For the latter, we further distinguish three sub-patterns: \textit{cascading} failures occur in close succession with substantial temporal overlap; \textit{rolling} failures are spaced further apart with partial overlap; \textit{independent} failures are fully separated with no overlap.
Disk failures are fixed by rebinding the target's device driver to its server.
The impact of disk failures can vary from minimal to high depending on the number of disks targeted; despite the compact topology, RING's distributed architecture is resilient to disk anomalies by design. Therefore, multi-disk failure scenarios produce stronger signals. 

\textbf{\textit{(ii) Server failures}} are triggered by tampering with the target server's network interface. \textit{Single} server failures drop all inbound traffic from the rest of the cluster, and are bound to a certain duration. 
This simulates a complete network partition, as would occur during a switch failure or a severe network misconfiguration. \textit{Asynchronous} failures target multiple servers at different timestamps, and, like disk failures, are categorized in \textit{cascading, rolling, independent}. \textit{Degradation} failures progressively increase latency and packet loss between the target server and the rest of the cluster, following either a \textit{linear} ramp or \textit{staged} discrete steps. This simulates progressive network deterioration, such as a slowly failing network interface or increasing link congestion. Server failures are fixed by reversing network changes and returning the server to its nominal state. Server failures have a strong impact on the rest of the cluster 
and thus heavily influence metrics across the entire system. Notice that  \textit{Simultaneous} server failures are excluded as they would essentially render the cluster non-functional.

\begin{figure}[t]
\centering
\includegraphics[width=1.0\textwidth]{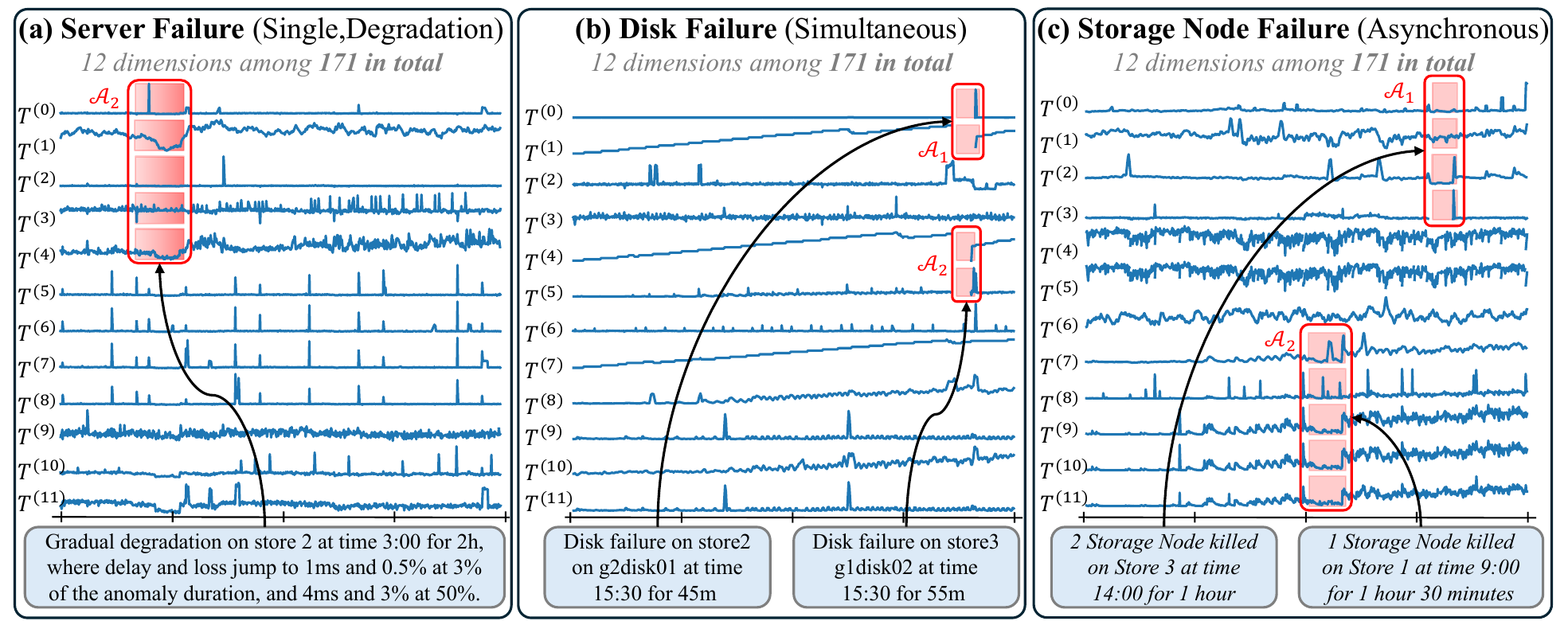}
\vspace{-0.2cm}
\caption{Sample of time series for each of the $3$ main anomaly types in \textbf{SHAD}.}
\vspace{-0.6cm}
\label{fig:anomalies_example}
\end{figure}

\textbf{\textit{(iii) Storage node failures}} are triggered by instantly killing the process(es) corresponding to a configurable number of software nodes on the targeted storage server. They simulate software-level crashes which, while relatively rare, can occur in edge case scenarios. Storage node failures have the same typology as disk failures, occurring as \textit{Single}, \textit{Simultaneous}, or \textit{Asynchronous} failures with \textit{cascading}, \textit{rolling}, and \textit{independent} sub-patterns.
Storage Node failures have an additional parameter corresponding to the number of random nodes (from 1 to 3) killed on each targeted server per event. Storage node failures are fixed by restarting the process corresponding to each failed node, after which the RING will run background tasks until the targeted servers are back in their nominal state.
The impact of storage node failures is typically the smallest out of the three types of anomalies, because the DHT efficiently redistributes the failed nodes' work across the remaining live nodes.

A visual example of the anomalies described above is shown in Figure \ref{fig:anomalies_example}, along with a textual explanation of the anomalies involved. 
Although our experiments were conducted on a specific system, the considered anomalies are not platform-specific. Failures such as disk loss, network partitions, and software crashes are common across distributed storage systems. 
Parameters for each anomaly, such as time, duration, and subsequent anomaly-specific variables, were chosen with the goal of ensuring broad coverage across timeline and failure configurations. Across all anomaly types, failure onset times span across the full 18-hour experiment (from 30 minutes to 16.5 hours); durations range from 1 minute to over 9 hours, and target components are evenly distributed across the 3 storage servers  (see Table~\ref{tab:coverage} in the Appendix).

\begin{wrapfigure}[17]{r}{0.45\columnwidth}
  \centering
  \vspace{-0.5cm}
  \includegraphics[width=\linewidth]{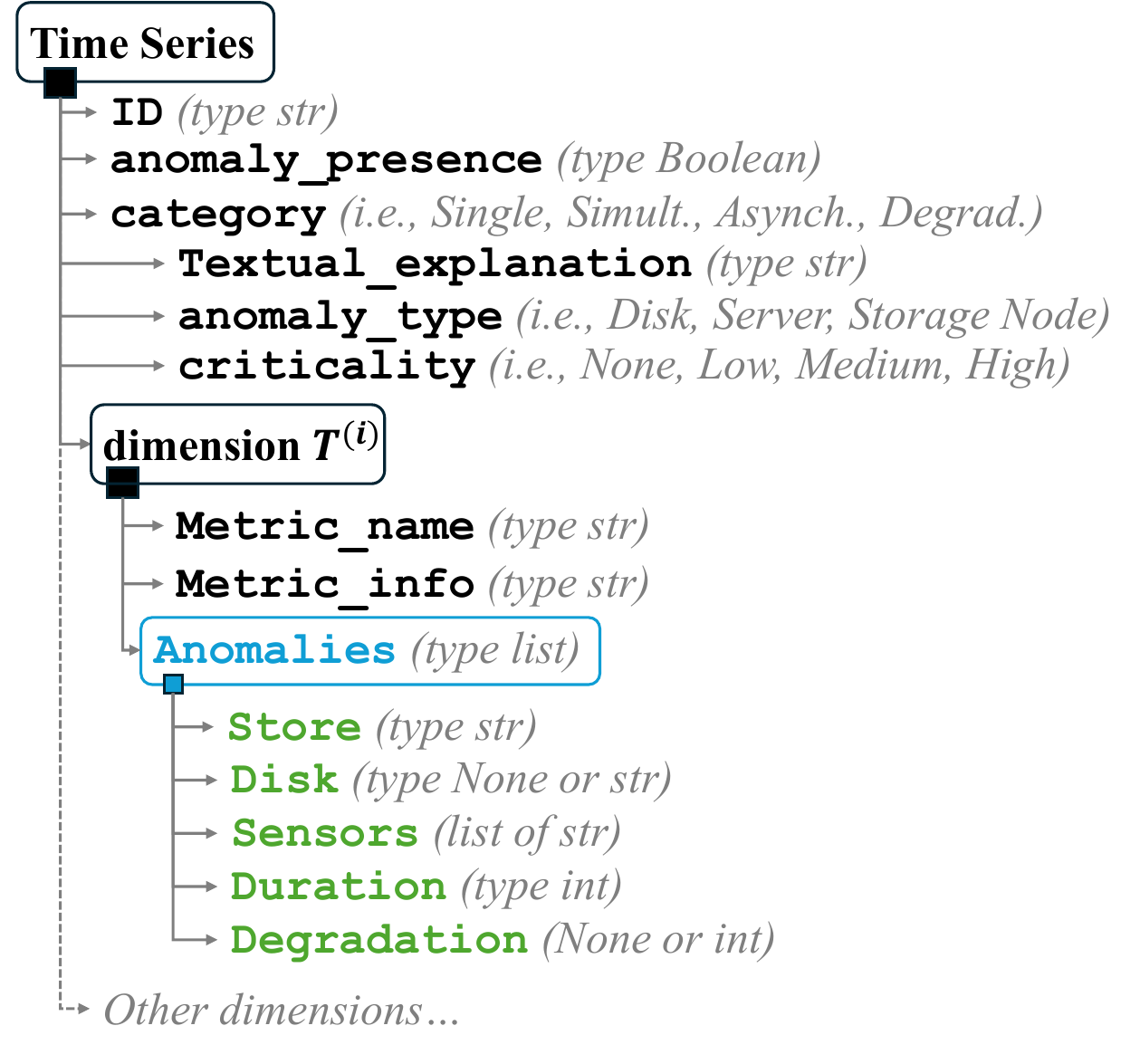}
  \caption{SHAD's annotations}    \label{fig:annotation_structure}
\end{wrapfigure}
\textbf{Dataset Characteristics.} SHAD contains 215 multivariate time series, whose composition is detailed in Figure~\ref{fig:shad_overview}. 
Each series contains ${\sim}1021$ data points, which were sampled with a 1-minute resolution across 171 features (listed in Appendix~\ref{app:metrics}). The anomalies are annotated by Scality experts, who supervised each planned experiment, ensuring that the injected failure type, timing, and affected components were accurately recorded and consistently reflected in the released metadata.  
Anomalies are labeled on two layers:
\textbf{Top-level labeling} assigns one binary value per time point, corresponding to whether an anomaly is ongoing and aligning with the format of existing benchmarks \cite{liu2024elephant}.
\textbf{Dimension-level} labeling similarly flags anomalies
per time point, but only to the dimensions corresponding to
the \textit{root cause} components they directly target.
In addition, to support interpretability and explainability studies, each time series in SHAD is accompanied by a \textbf{structured XML annotation} (see Figure~\ref{fig:annotation_structure}), with both global and fine-grained information. For each series, we provide whether an anomaly is present, its  criticality, and its category, along with a textual explanation. Then, fine-grained  per-dimension annotations include a textual explanation and, in case of anomalies, failure-specific parameters, also incorporating the set of other dimensions involved by the same anomaly, allowing for linking related anomalous behaviors and reconstructing the full system-level event.
To our knowledge, no existing TSAD benchmark provides comparable semantic annotations, while SHAD enables new forms of evaluation for both  explainability and interpretability.

\begin{figure}[tb]
\centering
\includegraphics[width=\textwidth]{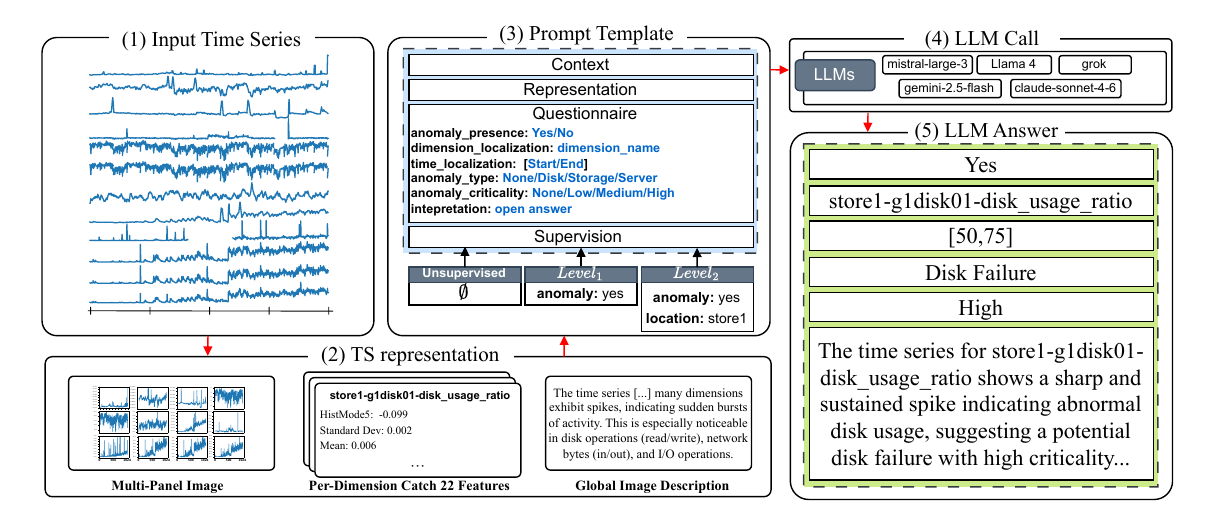}
\caption{Overview of the evaluation for interpretability.}
\vspace{-0.3cm}
\label{fig:llm_evaluation}
\end{figure}

\section{Experimental Evaluation on SHAD} \label{sec:experiments}
We now discuss the experiments aimed at addressing our research questions. For \textbf{RQ1} (\underline{\textit{Detection}}), we assess the effectiveness of 18 state-of-the-art detection methods covering a wide range of Time Series Anomaly Detection processes (see Table \ref{tab:detector_list} in Appendix \ref{app:experimental_evaluation} for the full list) on SHAD. Both Semi-Supervised (trained on nominal data) and Unsupervised approaches are included, using the optimized hyperparameters from the TSB-AD benchmark~\cite{liu2024elephant}. Performance is evaluated using VUS-PR (a robust, threshold-independent metric) with a 25-point buffer \cite{paparrizos2022volume,boniol2025vus}.
For \textbf{RQ2} (\underline{\textit{Explanation}}): we evaluate the contribution of each dimension in the final anomaly score for the top performing method of each supervision category. We evaluate their performance using the NDCG metric~\cite{wang2013ndcg}, since this metric rewards well ranked features diagnosis.
For \textbf{RQ3} (\underline{\textit{Interpretation}}), we evaluate the ability of frozen LLMs to interpret anomalies using a single call. Each time series is provided to the LLM under different representations: (i) \textit{Catch22} features \cite{lubba2019catch22} for each dimension; (ii) a \textit{multi-panel image} with one subplot per dimension; and (iii) a \textit{natural language description} of the time series image generated by the same LLM\footnote{We also considered ChatTS \cite{xie2024chatts} for descriptions, but it supports only up to $30$ time series at the time of writing.}.
The LLM is prompted with a structured questionnaire (Figure \ref{fig:llm_evaluation}) to: (i) detect anomaly presence, (ii) identify at least one affected dimension, (iii) localize the anomaly in time, (iv) classify it, and (v) assess its severity.
We consider \textit{F1-score} for classification tasks (presence, type, severity) and \textit{accuracy} for localization tasks, averaged over 5 runs. We compare five LLMs (listed in Figure \ref{fig:llm_evaluation}), and we also performed an additional evaluation with models from the same family (i.e., Mistral) with different capacities to evaluate whether the LLM size has an impact on interpretability. Finally, LLM experiments are repeated under two supervision levels to simulate an \textit{end-to-end} TSAD pipeline that covers all three stages (\textit{detection, explanation, interpretation}). In particular, we simulate the availability of an always correct anomaly detector (i.e., \textit{level$_1$}, which provides the LLM with information about anomaly presence), and the availability of an always correct anomaly explainer  (i.e., \textit{level$_2$}, which provides the LLM with information about anomaly presence \textit{and} the affected system components).

\subsection{RQ1: \textit{Time Series Anomaly \underline{Detection}}}
\label{sec:experiments_tsad}

\begin{figure}[t]
\centering
\includegraphics[width=\textwidth]{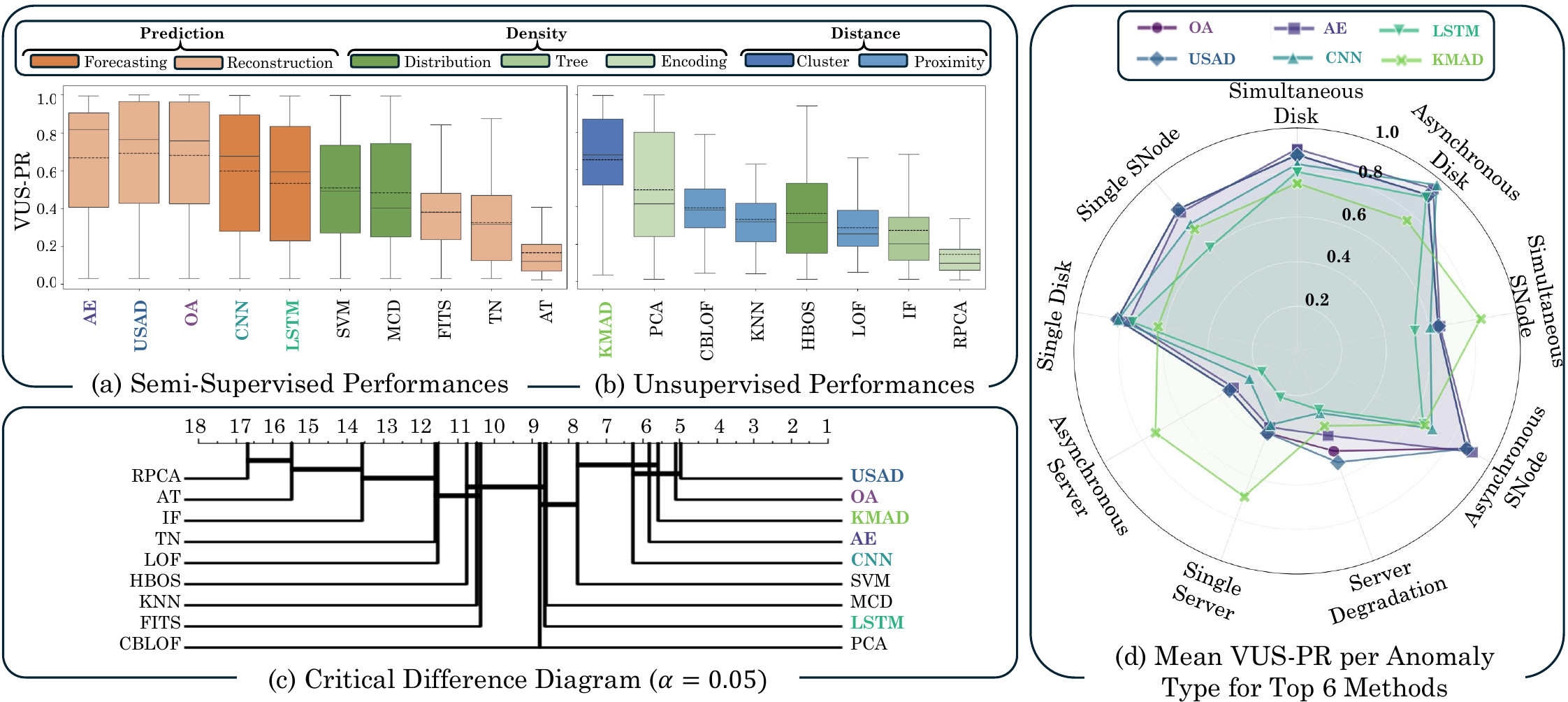}
\caption{Anomaly Detection on SHAD}
\vspace{-0.3cm}
\label{fig:Detection}
\end{figure}
Figure~\ref{fig:Detection} (a) and (b) present the VUS-PR performance of respectively Semi-Supervised and Unsupervised approaches on \textbf{SHAD}, ordered per median performance (full line).
A first observation is that the top Semi-Supervised methods significantly outperform the Unsupervised ones, with the notable exception of KMAD which is on par with the top-5 Semi-Supervised.
This is expected, as the Semi-Supervised methods were trained on the 83 nominal time series, providing a large enough training set for deep learning methods to outperform the more classical approaches, concordant with the conclusions of~\cite{audibert22DNN}.
The critical difference diagram in Figure~\ref{fig:Detection} (c) provides consistent observation, as the top ranked methods in median performance, highlighted in bold colors, belong to a primary clique indicating no statistically significant difference between them, despite their different learning paradigms. 
Specifically, the top-performing semi-supervised methods (OA, USAD, AE and CNN) are joined by the unsupervised KMAD under a single horizontal bar, confirming that KMAD is statistically competitive with the best semi-supervised approaches at $\alpha = 0.05$.
Figure~\ref{fig:Detection} (d) takes a closer look at the best performing methods, providing their mean performance on the different anomaly types existing in \textbf{SHAD}.
A first observation is that the behavior of these methods is consistent within their process-centric categorization: OA, AE and USAD, which are Reconstruction-based methods, have a different spider-plot than CNN and LSTM, which are Prediction-based. Furthermore, these Semi-Supervised methods have a significantly distinct performance than KMAD, which is Unsupervised.
More specifically, this figure highlights the different paradigms taken by each supervision modes. 
While the semi-supervised methods (OA, USAD, AE, CNN, LSTM) generally offer higher precision on subtle anomalies, the unsupervised KMAD exhibits superior robustness in detecting system-wide network events.
Indeed, "small impact" failures such as the Single Snode or Single Disk failures, as described in Section~\ref{sec:Benchmark}, produce faint signals that only the Semi-Supervised methods manage to detect, as these low effects can be overlooked as noise by Unsupervised approaches. However, this learning is detrimental when faced with system-wide events, such as Asynchronous server or Single Server failures. When those event occur, with a majority of the metrics shifting to never seen data, DNN models may get lost, whereas Unsupervised approaches, such as KMAD, that rely on a global view of the data, will detect such a clear shift happening.

\begin{tcolorbox}[colback=gray!5!white,colframe=gray!80!white,boxsep=1pt,top=1pt,bottom=1pt]
\small
\textit{These results show that \textbf{SHAD} benchmark remains non-trivial in Anomaly Detection, as no type of method manages to be exhaustively accurate on each anomaly type.}
\end{tcolorbox}

\subsection{RQ2: \textit{Time Series Anomaly \underline{Explanation}}}
\label{sec:experiments_exp}

\begin{wrapfigure}[12]{r}{0.5\columnwidth}
  \centering
  \vspace{-0.5cm}
  \includegraphics[width=\linewidth]{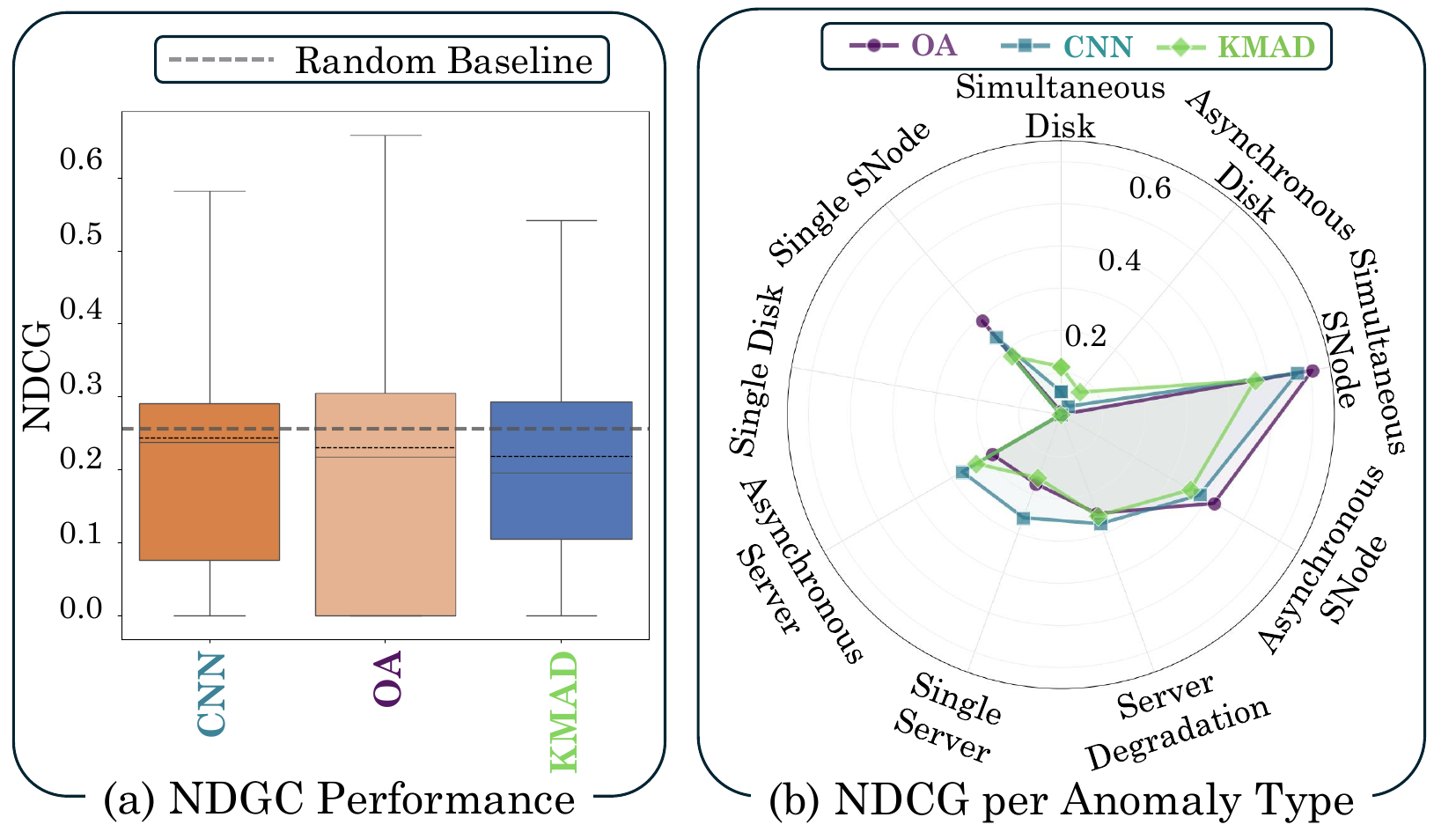}
  \vspace{-0.5cm}
  \caption{Basic Explainability Results}    \label{fig:Explainability}
\end{wrapfigure}

Beyond anomaly detection accuracy, we also evaluate the native capacity of the best performing methods to accurately localize the anomalies across the feature space. 
We select three representative methods: OA (Semi-Supervised, Reconstruction-based), CNN (Semi-Supervised, Prediction-based) and KMAD (Unsupervised, Clustering-based), and derive their feature-wise scores.
Results in Figure~\ref{fig:Explainability} (a) reveal a striking observation: despite their strong detection performance, the methods exhibit behavior nearly indistinguishable from a random baseline in term of dimension localization. 
This highlights that good detection accuracy does not inherently translate into diagnostic utility, and internal scoring mechanisms of these methods are not natively designed for feature attribution.
The granular analysis on the anomaly type in Figure~\ref{fig:Explainability} (b) confirms this observation.
The worst results (Single Disk, Simultaneous Disk and Asynchronous Disk) have a low dimensional contamination rate (respectively 4, 12 and 12 dimensions contaminated on average) while the best localized anomalies, Simultaneous SNode, have on average 140 anomalous dimensions.

\begin{tcolorbox}[colback=gray!5!white,colframe=gray!80!white,boxsep=1pt,top=1pt,bottom=1pt]
\small
\textit{The inherent diversity and complexity of \textbf{SHAD} establish it as a non-trivial explainability testing ground that cannot be addressed with simple dimension contribution analysis.}
\end{tcolorbox}

\begin{figure}[t]
\centering
\includegraphics[width=1.0\textwidth]{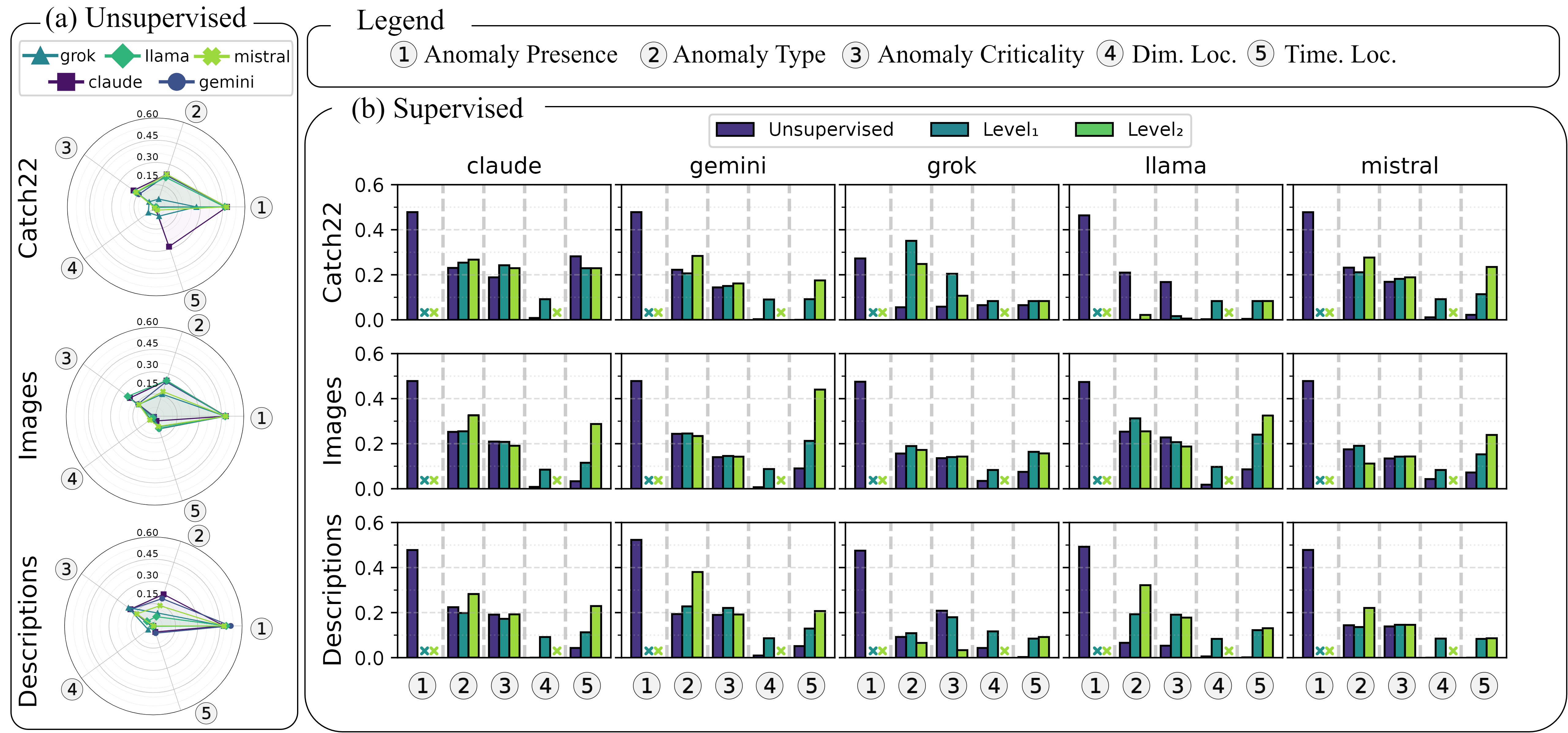}
\caption{Results for the interpretability experiment. For \textit{Presence}, \textit{Type}, and \textit{Criticality} we consider the \textit{F1-score}, while for \textit{Dimension} and \textit{Time Localization} we report the \textit{accuracy}. } 
\vspace{-0.5cm}
\label{fig:all_llm_results_supervision}
\end{figure}

\subsection{RQ3: \textit{Time Series Anomaly \underline{Interpretation}}}
\label{sec:experiments_int}
\textbf{Comparison with different representations.} 
Figure \ref{fig:all_llm_results_supervision}(a) reports the performance of LLMs across different input representations for the interpretability task. Overall, results are very low, highlighting the difficulty of interpreting high-dimensional real-world time series. For \textit{anomaly presence} (\graycircle{1}), most of the LLMs always predicted an anomaly regardless of the representation, suggesting that irregular patterns and high-dimensionality lead the models to interpret variability as anomalies. This led to very similar average results across all models (0.48, 0.47, 0.43 for Descriptions, Images, and Catch22, resp.). For \textit{anomaly type}, performance remains poor, with  the best \textit{F1-score} from Images ($21\%$), followed by Catch22 ($19\%$) and descriptions ($14\%$), and a noticeable bias toward Disk Failures. The \textit{anomaly criticality} task (\graycircle{3}) is even more challenging, with a maximum of $16\%$ (Images), showing the inability to assess severity without contextual knowledge. \textit{Dimension localization} (\graycircle{4}) is among the hardest tasks, with near-zero accuracy ($1\%$, $2\%$, $1\%$ for Catch22, Images, and descriptions), reflecting the difficulty of identifying anomalous dimensions in large multivariate data. For \textit{time localization} (\graycircle{5}), results are slightly better but still low ($7\%$, $7\%$, $2\%$ for Images, Catch22, and Descriptions). Overall, all the  results highlighted the complexity of interpreting time series using LLMs without any type of supervision. While the results are generally low, providing Images was the best option for three tasks (\graycircle{2}, \graycircle{3}, \graycircle{4}) and the second best option for the others, suggesting that visual representations provide a slightly clearer overview of the series.

\textbf{Comparison with different LLMs.} 
Figure \ref{fig:all_llm_results_supervision}(a) compares the performance of 5 LLMs for each representation across the five tasks. For \textit{anomaly presence} (\graycircle{1}), almost all models behaved identically ($\approx47\%$ across all representations), showing a strong bias towards anomalies. The only exception was Grok (especially with  the Catch22 representation) which, in contrast, judged most series as normal. This is consistent with recent evidence that Grok exhibits a distinctive behavioral profile compared to other frontier LLMs \cite{zhang2025stress}. 
For \textit{anomaly type} and \textit{criticality} (\graycircle{2}, \graycircle{3}) the best LLM is Claude ($0.23$ and $0.19$, resp.), while Grok was the best for \textit{dimension localization} (\graycircle{4}), but with only 4\% of accuracy. Finally, Claude was the best model for \textit{time localization} (\graycircle{5}), with 12\%. Overall, all the models showed very similar behaviours (we provide the full results in Appendix \ref{app:experimental_evaluation}), showing that interpreting anomalies is a significant challenge.

\textbf{Adding supervison.} We evaluate LLM performance under two supervision levels, simulating the \textit{interpretability} step when \textit{detection} and \textit{explanation} outputs are available. Figure \ref{fig:all_llm_results_supervision}(b) reports results across all models and representations. For clarity, we omit tasks directly answered by supervision (accuracy $=1.0$).
With \textit{level$_1$} (i.e., anomaly presence), performance consistently improves for \textit{dimension} and \textit{time localization} (\graycircle{4}, \graycircle{5}), with average gains of $7.2\%$ and $7.9\%$. Although results remain low, this indicates that additional information can help characterize anomalies. For other tasks, trends are inconsistent, with negligible variation (below $0.2\%$). With \textit{level$_2$} (i.e., anomaly presence + affected components), further improvements are observed for \textit{time localization} (\graycircle{5}) (+$14.4\%$) and \textit{anomaly type} (\graycircle{2}) (+$4.7\%$), while \textit{anomaly criticality} shows negligible change. Overall, although combining \textit{detection} and \textit{explanation} signals improves performance, zero-shot LLMs remain unable to reliably characterize anomalies, highlighting the need for domain-specific solutions.

\begin{tcolorbox}[colback=gray!5!white,colframe=gray!80!white,boxsep=1pt,top=1pt,bottom=1pt]
\small
\textit{Zero-shot, single-call LLMs cannot properly interpret SHAD's anomalies. Adding information from \textbf{detection} and \textbf{explanation} provides some benefits, but challenges still remain, suggesting the need for tailored, domain-specific solutions. }
\end{tcolorbox}

\begin{figure}[t]
\centering
\includegraphics[width=0.9\textwidth]{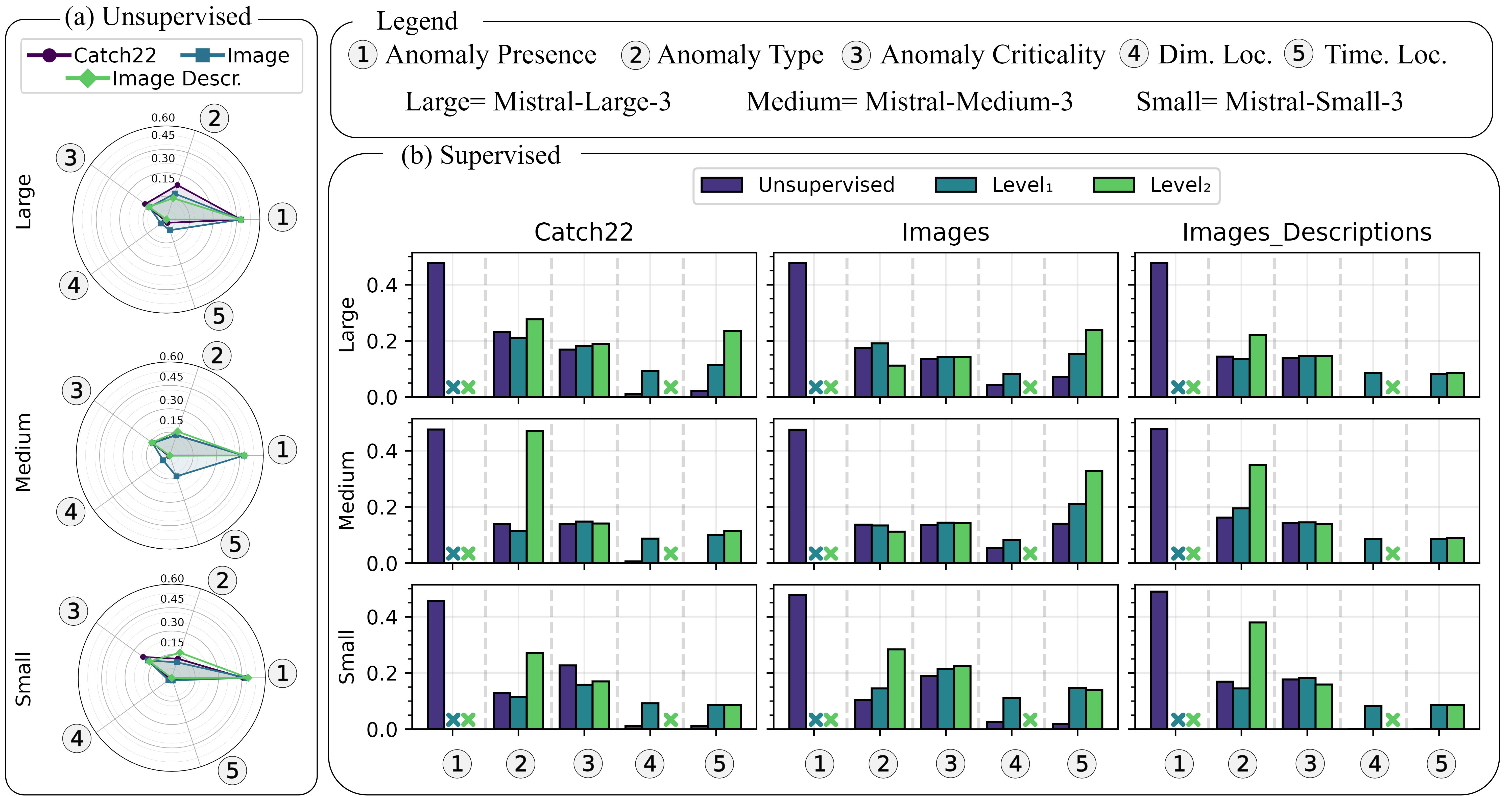}
\caption{Interpretability results with the same model family (Mistral large/medium/small).}
\vspace{-0.5cm}
\label{fig:llm_results_supervision}
\end{figure}

\textbf{Comparison with smaller LLMs.} 
Figure \ref{fig:llm_results_supervision}(a) compares the performance of Mistral Large, Medium, and Small. With Mistral Medium, we observe a significant drop when using Catch22, likely due to the difficulty of reasoning over many statistical features across high-dimensional data. In contrast, performance with Images remains stable and even improves on \textit{time localization} (\graycircle{5}) ($14\%$ vs. $7\%$). Results with Image Descriptions are nearly identical to the largest model. With Mistral Small, performance  generally drops, highlighting its limited capacity to handle both high-dimensional statistical features and richer multimodal inputs, leading to consistently lower results across tasks.

\section{Conclusion and Future directions}\label{sec:Conclusions}
We proposed \textbf{SHAD}, a novel end-to-end benchmark composed of real-world, high-dimensional time series, that covers all the steps in TSAD pipelines (\textit{detection, explanation}, and \textit{interpretation}). 
SHAD includes both normal and anomalous series, with anomalies affecting different components and severity levels, and provides per-dimension labels along with rich textual annotations for each dimension and anomaly.
Our experimental evaluation concluded that (i) SOTA anomaly detection approaches are accurate but not perfect for all anomalies in SHAD, (ii) Explanability on SHAD cannot be natively solved with existing TSAD methods, and (iii) frozen LLMs are not able to provide correct interpretations of anomalies. 
Finally, SHAD enables new research directions: in \textit{detection}, studying robustness when directly affected metrics are removed; in \textit{explainability}, rigorously assessing variable contributions via per-dimension labels and developing dedicated explainability methods trained on SHAD's annotations, rather than relying on native detector attribution scores; and in \textit{interpretability}, investigating how reliable dimension-level explanations can be provided as structured, semantically enriched context to LLMs, enabling the generation of human-understandable anomaly interpretations grounded in both data and semantics.

\begin{ack}
The authors are grateful to the CLEPS infrastructure from the Inria of Paris for providing resources and support.
\end{ack}

\bibliographystyle{abbrvnat}
\bibliography{references}

\include{appendix}


\end{document}

%% file: appendix.tex
\clearpage
\appendix
\appendixpage
\addappheadtotoc
\startcontents[sections]
\printcontents[sections]{l}{1}{\setcounter{tocdepth}{2}}

\clearpage

\section{Related Work}
\label{app:related}

\subsection{Existing benchmarks}
\paragraph{Existing Benchmarks: \textit{How do we evaluate time series anomaly detection?}} 
Research in TSAD has led to the introduction of numerous datasets (see Table ~\ref{tab:datasets}), which have been considered in multiple  surveys~\cite{paparrizos2022tsb, schmidl2022anomaly,sylligardos2023choose,liu2024elephant}, consistently highlighting the absence of a one-size-fits-all anomaly detector. Consequently, efforts have aimed to establish benchmarks, whose main characteristics are detailed in Table \ref{tab:benchmark_comparison}, incorporating multiple datasets from various domains for comprehensive evaluation.
\textbf{NAB}~\cite{lavin2015evaluating} provides 58 labeled real-world and artificial time series, and 
primarily focuses on real-time anomaly detection for streaming data.
It comprises diverse domains such as AWS server metrics, online advertisement clicking rates, real-time traffic data, and a collection of Twitter mentions of large publicly-traded companies.
 \textbf{TODS}~\cite{lai2021revisiting} refines synthetic criterion and includes five anomaly scenarios categorized as point-global, pattern-contextual, pattern-shapelet, pattern-seasonal, and pattern-trend. \textbf{TimeEval}~\cite{schmidl2022anomaly} comprises a collection of datasets (both real/synthetic and univariate/multivariate) with point and sequence anomalies. 
 In addition, this benchmark has been filtered such that there is no time series that has a normal/abnormal ratio above $0.1$, and that at least one method performs more than 0.8 AUC-ROC for each time series.
\textbf{TimeSeAD}~\cite{wagner2023timesead} analyzed widely used multivariate datasets and introduced a recall-consistent evaluation metric that accounts for temporal correlations, benchmarking 28 deep-learning based methods, while \textbf{Experimental-TSAD}~\cite{zhang2023experimental} compared 17 state-of-the-art anomaly detection methods using both point-based and range-based metrics across various data sizes, anomaly rates, feature dimensions, and temporal patterns. 
\textbf{TSB-UAD}~\cite{paparrizos2022tsb} is a comprehensive benchmark designed for evaluating univariate TSAD, including real-world and synthetic datasets.
\textbf{Exathlon}~\cite{jacob2020exathlon} is proposed for explainable anomaly detection over high-dimensional time series data. It is constructed based on real data traces from an Apache Spark cluster, and was designed to address the pitfalls of previous benchmarks~\cite{wu2021current}. It features a single anomalies per series, which in most cases is artificially.
Aiming at addressing the challenge of designing truly representative benchmarks, \textbf{TSB-AD}~\cite{liu2024elephant} introduced the first large-scale, heterogeneous, and meticulously curated dataset, also incorporating latest advances in foundation model-based methods, highlighting their potential for TSAD.
Finally,  \textbf{TAB}~\cite{qiu2025tab} is a recent and comprehensive benchmark that provides an extensive algorithmic comparison, including 19 statistical methods, 18 neural network–based approaches, and 7 foundation model–based methods. 

\subsection{TSAD Datasets}
\label{app:datasets}
Table \ref{tab:datasets} lists the datasets for time series anomaly detection currently available in the literature. For each one, we report the key characteristics, including the number of available time series and their average lenght and dimensionality, the type of anomalies, and the domain from which they belong. 

\subsection{Interpretability and Explainability}
\label{app:int_exp}

\paragraph{Interpretable and Explainable Anomaly Detection.}
Explainable machine learning techniques can be broadly categorized into \textit{intrinsically explainable models} and 
\textit{model-agnostic} approaches~\cite{jacob2020exathlon}. The former category focuses on building a human-readable representation (e.g. \textit{decision trees}) from data \cite{molnar2020interpretable}, while the latter separate explanations from the ML model~\cite{ribeiro2016should}. From the perspective of anomaly detection approaches, all forecasting/reconstruction methods~\cite{Wu_2022,Omnianomaly} are inherently explainable. Among model-agnostic approaches, several methods are based on perturbing the inputs and observing the response of the classifier~\cite{lakkaraju2016interpretable}. LIME~\cite{ribeiro2016should} explains individual predictions by learning a local surrogate model that approximates the original model in the neighborhood of a given instance, using a weighted linear regression formulation. This framework was extended by Anchors~\cite{ribeiro2018anchors}, which replaces the linear model with logical rules. Other approaches assign a numerical score to each feature~\cite{bertossi2020causality, sundararajan2017axiomatic}, indicating its importance in the classification of each instance. Among these, SHAP~\cite{lundberg2017unified} provides a theoretically grounded measure of feature importance based on Shapley values \cite{winter2002shapley},
and has been recently extended to support time series data ~\cite{nayebi2023windowshap}.
Other approaches focus on providing \textit{global explanations}, approximating deep-learning classifiers with explainable models~\cite{frosst2017distilling, dong2024interpretability} or by learning decision rules~\cite{kopp2020anomaly, lakkaraju2016interpretable, chen2025novel}.

In the context of anomaly detection, the literature on interpretable approaches is rather limited. Among existing methods, MacroBase \cite{bailis2017macrobase} supports the user by prioritizing their attention on specific portions of high-volume data streams and by providing interpretable insights based on the relative recurrence of key attributes. On the other hand, EXstream \cite{zhang2017exstream} finds interpretations to distinguish the normal and anomalous periods given as input by the user. 
Recent advances in large language models (LLMs) represent a promising research direction in this area, enabling the generation of context-rich insights by leveraging model outputs and/or auxiliary information \cite{jin2024position}. While many approaches focused on using LLMs for the anomaly detection task itself \cite{jin2023time,wang2024news,rasul2023lag}, other have started to explore their usefulness in providing human-understandable data analysis \cite{xing2021deepsqa, oh2023ecg}. 
In particular, recent years have witnessed the introduction of several Time Series Question Answering (TSQA) systems in domains such as healthcare and sensor analytics \cite{xing2021deepsqa, oh2023ecg}, where natural language interfaces are used to query and interpret raw temporal signals. 
LLMs have also been explored for integrating time series with external textual context, enabling more informed predictions and decision-making \cite{kurisinkel2024text2timeseries, lai2025llmlight}. Agent-based and tool-augmented frameworks further enhance autonomous reasoning and support root cause analysis \cite{wang2024rcagent, zhou2023d}. While several LLM-based approaches have been proposed \cite{gruver2023large, zhou2023one}, only few of them supports multimodal data \cite{zhuang2024see, xie2024chatts}. Among these, ChatTS~\cite{xie2024chatts} is a recent multimodal LLM (MLLM) that can perform both reasoning and understanding on time series. It directly takes multivariate time series as input and enables natural language interaction for interpreting patterns, relationships, and anomalies within the data.

\begin{table}[t]
  \scriptsize
  \centering
  \caption{Summary characteristics of 30 datasets}
  \resizebox{\columnwidth}{!}{
  {\tiny
\begin{tabular}{l|c|c|c|c|c|c}
\multicolumn{1}{l|}{\textbf{Name}} & \textbf{Category} & \textbf{Dimension} & \textbf{Avg Dim} & \textbf{\# of time series} & \textbf{Avg Length} & \textbf{Domain}\\ \hline
UCR~\cite{wu2021current} & P\&S & I & 1 & 250 & 67818.7 & Multiple\\
Yahoo~\cite{laptev2015s5} & P\&S & I & 1 & 367 & 1560.2 & Web Service\\
IOPS~\cite{IOPS} & S & I & 1 & 58 & 72792.3 & IT/Web Services\\ 
MGAB~\cite{thill2020markusthill} & S & I & 1 & 10 & 97777.8 & Synthetic\\ 
WSD~\cite{zhang2022efficient} & S & I & 1 & 210 & 17444.5 & IT/Web Services\\ 
SED~\cite{boniol2021sand} & S & I & 1 & 6 & 23332.3 & Sensor\\ 
NEK~\cite{si2024timeseriesbench} & P\&S & I & 1 & 48 & 1073.0 & IT/Web Services\\ 
Stock~\cite{tran2016distance} & P\&S & I & 1 & 90 & 15000.0 & Financial\\ 
Power~\cite{keogh2007finding} & S & I & 1 & 1 & 35040.0 & Facility\\ 
SensorScope~\cite{yao2010online} & S & I & 1 & 23 & 27038.4 & Environmental\\
Dodgers~\cite{10.1145/1150402.1150428} & S & I & 1 & 1 & 50400.0 & Traffic\\

Daphnet~\cite{bachlin2009wearable} & S & M & 9 & 17 & 38774.0 & Human Activity\\ 
CATSv2~\cite{fleith2023controlled} & S & M & 17 & 10 & 240000.0 & Sensor\\ 
SWaT~\cite{goh2016dataset} & S & M & 59 & 4 & 207457.5 & Sensor\\ 
LTDB~\cite{goldberger2000physiobank} & S & M & 2 & 7 & 100000.0 & Medical\\ 
TAO~\cite{tao_dataset} & P\&S & M & 3 & 45 & 10000.0 & Aerospace\\ 
MITDB~\cite{goldberger2000physiobank} & S & M & 2 & 48 & 336153.8 & Medical\\ 
MSL~\cite{hundman2018detecting} & S & M & 55 & 27 & 3119.4 & Aerospace\\ 
SMAP~\cite{hundman2018detecting} & S & M & 25 & 54 & 7855.9 & Aerospace\\ 
SMD~\cite{su2019robust} & S & M & 38 & 28 & 25466.4 & IT/Web Services\\ 
SVDB~\cite{greenwald1990improved} & S & M & 2 & 78 & 207122.6 & Medical\\ 
OPP~\cite{roggen2010collecting} & S & M & 248 & 24 & 17426.75 & Human Activity\\
GHL~\cite{filonov2016multivariate} & S & M & 19 & 48 & 199001.0 & Environmental\\ 
Genesis~\cite{von2018anomaly} & S & M & 18 & 1 & 16220.0 & Sensor\\ 
PSM~\cite{abdulaal2021practical} & P\&S & M & 25 & 1 & 217624.0 & Facility\\ 
CreditCard~\cite{sharafaldin2018toward} & P\&S & M & 29 & 1 & 284807.0 & Financial\\ 
GECCO~\cite{moritz2018gecco} & S & M & 9 & 1 & 138521.0 & Sensor\\ 
SMD~\cite{su2019robust} & S & M & 38 & 28 & 25300.7 & Facility\\
Occupancy~\cite{candanedo2016accurate} & S & M & 5 & 2 & 6208.5 & Environmental\\
Metro~\cite{metro} & S & M & 5 & 1 & 48204.0 & Traffic\\
\hline
\end{tabular}
}}
\begin{tablenotes}
    \centering
    \small
    \item  I: Univariate; M: Multivariate // P: Point; S: Subsequence
    \item The statistics are based on the datasets downloaded during the writing of this article.
\end{tablenotes}
\vspace{0.2cm}
\label{tab:datasets}
\end{table}

\newpage

\section{Metric Catalog}
  \label{app:metrics}
Tables~\ref{tab:metrics_platform}--\ref{tab:metrics_device} provide the complete catalog of all 171 dimensions in the dataset, grouped by hierarchy level. 
Server-level metrics (Table~\ref{tab:metrics_server}) are replicated across all three storage servers with the naming convention \texttt{\{server\}-\{metric\}} (e.g., \texttt{store1-cpu\_usage\_ratio}), yielding $19 \times 3 = 57$ columns. 
Device-level metrics (Table~\ref{tab:metrics_device}) follow \texttt{\{server\}-\{device\}-\{metric\}} (e.g., \texttt{store1-g1disk01-disk\_read\_bytes}), with 7 devices per server and 4 metrics per device, yielding $4 \times 7 \times 3 = 84$ columns.

Platform-level metrics include telemetry from the supervisor server, which also hosts the experiment framework. Its contribution is negligible and consistent across all experiments.

  \begin{table}[H]
    \scriptsize
    \centering
    \caption{Platform-level metrics (17 dimensions, cluster-wide aggregates).}
    \label{tab:metrics_platform}
    \begin{tabularx}{\textwidth}{l l X}
      \toprule
      \textbf{Metric} & \textbf{Unit} & \textbf{Description} \\
      \midrule
      cpu\_usage\_ratio & ratio [0,1] & Fraction of CPU time not idle, averaged across all servers. \\
      cpu\_load\_ratio & ratio [$\geq$0] & CPU load average normalized by CPU count across all servers. Can exceed 1.0 when the run queue exceeds CPU count. \\
      memory\_usage\_ratio & ratio [0,1] & Fraction of total used memory, averaged across all servers. \\
      network\_in\_bytes & bytes/s & Total incoming network throughput across all interfaces on all servers. \\
      network\_out\_bytes & bytes/s & Total outgoing network throughput across all interfaces on all servers. \\
      objects\_count & count & Number of objects stored on the RING, including replicas. \\
      objects\_unique\_count & count & Number of unique logical objects, deduplicated regardless of replica count. \\
      objects\_missing\_count & count & Number of objects with missing or out-of-service chunks. \\
      capacity\_disk\_unique\_tb & TB & Total stored data of unique objects before replication (logical data size). \\
      capacity\_disk\_stored\_tb & TB & Total stored data including all replicas. \\
      capacity\_disk\_used\_tb & TB & Actual disk space consumed, including storage overhead beyond object data. \\
      capacity\_disk\_available\_tb & TB & Free disk space on the RING. \\
      capacity\_disk\_total\_tb & TB & Total physical disk capacity. Static unless disks are removed. \\
      disk\_read\_bytes & bytes/s & Total disk read throughput summed across all disks. \\
      disk\_write\_bytes & bytes/s & Total disk write throughput summed across all disks. \\
      disk\_io\_time\_seconds & s/s [0,1] & Fraction of time spent on I/O, averaged across all disks. \\
      disk\_errors\_count & count & Number of disk errors across all disks. Always zero in this dataset. \\
      \bottomrule
    \end{tabularx}
  \end{table}

  \begin{table}[H]
    \scriptsize
    \centering
    \caption{S3-level metrics (13 dimensions, aggregated across storage servers).}
    \label{tab:metrics_s3}
    \begin{tabularx}{\textwidth}{l l X}
      \toprule
      \textbf{Metric} & \textbf{Unit} & \textbf{Description} \\
      \midrule
      s3\_errors\_5xx & errors/s & Rate of HTTP 5xx server errors across all S3 services. \\
      s3\_request\_get\_operations & ops/s & GET request rate. \\
      s3\_request\_post\_operations & ops/s & POST request rate. Near-constant (no multipart uploads in the nominal workload). \\
      s3\_request\_put\_operations & ops/s & PUT request rate. \\
      s3\_request\_delete\_operations & ops/s & DELETE request rate. \\
      s3\_request\_head\_operations & ops/s & HEAD (metadata lookup) request rate. \\
      s3\_throughput\_in\_bytes & bytes/s & Client-to-S3 data rate. \\
      s3\_throughput\_out\_bytes & bytes/s & S3-to-client data rate. Dominates throughput\_in in the GET-heavy workload. \\
      s3\_request\_get\_latency & seconds & 95th percentile GET request duration. \\
      s3\_request\_post\_latency & seconds & 95th percentile POST request duration. Near-constant. \\
      s3\_request\_put\_latency & seconds & 95th percentile PUT request duration. \\
      s3\_request\_delete\_latency & seconds & 95th percentile DELETE request duration. \\
      s3\_request\_head\_latency & seconds & 95th percentile HEAD request duration. \\
      \bottomrule
    \end{tabularx}
  \end{table}

  \begin{table}[H]
    \scriptsize
    \centering
    \caption{Server-level metrics (19 unique metrics $\times$ 3 servers = 57
    dimensions). Column names follow the pattern
    \texttt{store\{1,2,3\}-\{metric\}}.}
    \label{tab:metrics_server}
    \begin{tabularx}{\textwidth}{l l X}
      \toprule
      \textbf{Metric} & \textbf{Unit} & \textbf{Description} \\
      \midrule
      cpu\_usage\_ratio & ratio [0,1] & Fraction of CPU time not idle for this server. \\
      cpu\_load\_ratio & ratio [$\geq$0] & CPU load average normalized by CPU count. Can exceed 1.0. \\
      memory\_usage\_ratio & ratio [0,1] & Fraction of total used memory. \\
      network\_in\_bytes & bytes/s & All incoming network throughput (S3, RING inter-node, monitoring). \\
      network\_out\_bytes & bytes/s & All outgoing network throughput. \\
      operations\_sent & ops/s & Rate of internal RING operations sent to other servers. \\
      operations\_received & ops/s & Rate of internal RING operations received from other servers. \\
      latency\_sent\_milliseconds & ms & Average round-trip time for outbound RING operations. \\
      latency\_received\_milliseconds & ms & Average processing time for inbound RING operations. \\
      throughput\_sent\_bytes & bytes/s & Data payload sent via RING operations. \\
      throughput\_received\_bytes & bytes/s & Data payload received via RING operations. \\
      errors\_sent & errors/s & Rate of failed outbound RING operations. \\
      errors\_received & errors/s & Rate of failed inbound RING operations. \\
      rebuild\_scanned\_tasks & tasks/s & Rate of rebuild scanning (reconstructs missing replicas). \\
      rebuild\_rebuilt\_tasks & tasks/s & Rate of chunks actually rebuilt. \\
      repair\_scanned\_tasks & tasks/s & Rate of repair scanning (fixes integrity issues). \\
      repair\_repaired\_tasks & tasks/s & Rate of chunks actually repaired. \\
      balance\_scanned\_tasks & tasks/s & Rate of balance scanning (redistributes data for even usage). \\
      balance\_balanced\_tasks & tasks/s & Rate of chunks actually rebalanced. \\
      \bottomrule
    \end{tabularx}
  \end{table}

  \begin{table}[H]
    \scriptsize
    \centering
    \caption{Device-level metrics (4 metrics $\times$ 7 devices $\times$ 3
    servers = 84 dimensions). Column names follow
    \texttt{store\{1,2,3\}-\{device\}-\{metric\}}. Each server has 4 NVMe data
    disks (\texttt{g1disk01}, \texttt{g1disk02}, \texttt{g2disk01},
    \texttt{g2disk02}), 2 metadata SSDs (\texttt{ssd01}, \texttt{ssd02}), and 1
    OS root disk (\texttt{root}). Data disks are the targets of disk failure
    anomalies.}
    \label{tab:metrics_device}
    \begin{tabularx}{\textwidth}{l l X}
      \toprule
      \textbf{Metric} & \textbf{Unit} & \textbf{Description} \\
      \midrule
      disk\_read\_bytes & bytes/s & Disk read throughput. \\
      disk\_write\_bytes & bytes/s & Disk write throughput. \\
      disk\_io\_time\_seconds & s/s [0,1] & Fraction of time spent on I/O. Approaching 1.0 implies saturation. \\
      disk\_usage\_ratio & ratio [0,1] & Disk space utilization. \\
      \bottomrule
    \end{tabularx}
  \end{table}

\begin{figure}[h!]
\centering
\includegraphics[width=0.9\textwidth]{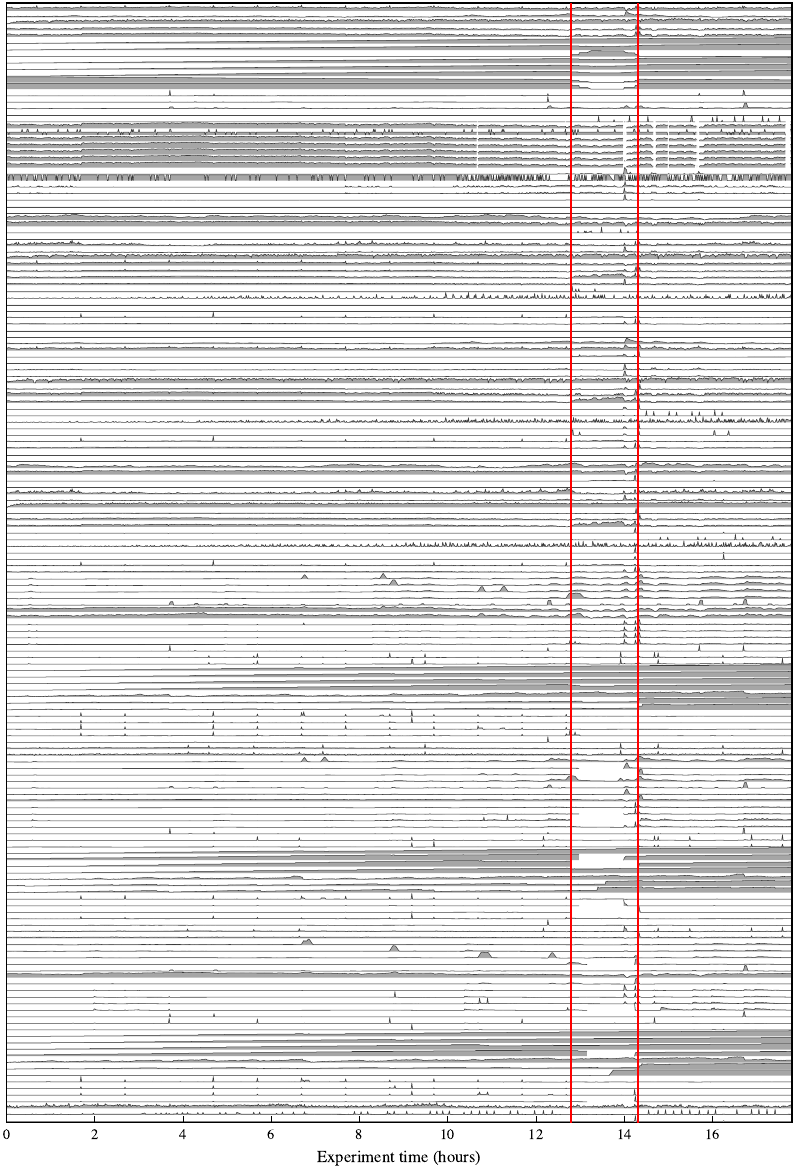}
\caption{Full 171 dimensions of an asynchronous, cascading disk failure experiment. Red vertical lines denote the full anomalous event, from the first disk failure onset to the last completed fix.}
\label{fig:full_exp}
\end{figure}

Figure~\ref{fig:full_exp} displays all 171 dimensions for a specific disk failure experiments (id=1773767292) with the anomaly period highlighted.

\newpage
\section{Experiment Registry}
\label{app:experiments}

In this section we provide the complete registry of all 215 experiments in the dataset, grouped by anomaly type. The \textbf{Details} column summarizes key anomaly configuration parameters (onset time, duration, target components, etc.). 
The \textbf{Notes} column contains additional comments on the experiment, including any experimental artifacts that occured. Namely, \textit{permanent disk loss} indicates the target disk could not be successfully recovered and stayed offline for the remainder of the experiment; \textit{server metric loss} indicates issues with the target's server telemetry due to software crashes; \textit{unplanned crash} indicates the target server could not be restarted automatically at anomaly remediation time.
\textit{rerun of ~ID} indicates the experiment is a rerun of an experiment containing artifacts; the resulting time series from such experiment and its rerun are therefore very different.

Experiment details use a compact notation summarized below.
\begin{itemize}

\item \textbf{Disk references} follow the pattern
  \texttt{s\textit{N}/\textit{disk}}, where \texttt{s\textit{N}} is the storage server number (1--3) and \textit{disk} is the NVMe device name (e.g.\ \texttt{g2disk01}).  Multiple disks are joined with~\texttt{+}.

\item \textbf{Timing} is given as \texttt{onset}~(time from
  experiment start to fault injection) and \texttt{dur}~(fault
  duration).  When every disk or node shares the same onset and
  duration, these appear once at the end of the entry; when they
  differ, the compact form \texttt{@\textit{onset}/\textit{dur}} is appended to each individual fault.  Faults within a non-simultaneous entry are separated by semicolons.

\item \textbf{Asynchronous patterns} are prefixed with one of
  three labels: \textit{cascading}~(failures start close together and mostly overlap), \textit{rolling}~(sequential with partial overlap), or \textit{independent}~(no temporal overlap).

\item \textbf{Server degradation} entries list either
  \texttt{max \textit{delay} / \textit{loss}} for linear ramps, or a sequence of \texttt{\textit{delay}/\textit{loss} @\textit{pct}\%} stages for gradual (staircase) profiles.  Delay is the added network latency; loss is the packet-drop percentage.  Stage percentages refer to elapsed fractions of the full anomaly duration.

\item \textbf{Software node (snode)} counts are written as
  \texttt{s\textit{N}: \textit{k} snode(s)}, indicating \textit{k}
  software storage nodes killed on server~\textit{N}.  When durations differ per server, they appear inline after the snode count.

\end{itemize}

\newcommand{\expid}[1]{\hypertarget{exp:#1}{\texttt{#1}}}
\newcommand{\expref}[1]{\hyperlink{exp:#1}{\texttt{#1}}}

\subsection{Baseline Experiments}
The nominal baseline is an 18\,h GET-heavy mixed-traffic workload
generated by \texttt{warp} with 16 concurrent workers, 15\,000 initial S3 objects of up to 100\,KiB (randomized), and a fixed operation distribution (GET~95\%, PUT~0.8\%, STAT~4\%, DELETE~0.4\%). All anomalous experiments listed in subsequent tables run on top of this same nominal workload.

\begin{scriptsize}
\begin{longtable}{l p{8cm} p{3cm}}
\caption{Nominal workload experiments (83 runs, no injected anomalies).}
\label{tab:registry-nominal} \\
\toprule
\textbf{ID} & \textbf{Details} & \textbf{Notes} \\
\midrule
\endfirsthead
\toprule
\textbf{ID} & \textbf{Details} & \textbf{Notes} \\
\midrule
\endhead
\bottomrule
\endfoot
\expid{1773251393} & 18h, 16 workers, 15k objects $\le$100KiB, GET 95\% / PUT 0.8\% / STAT 4\% / DEL 0.4\% & \\
\expid{1773251399} & \textrm{---\,\textit{ibid}\,---} & \\
\expid{1773343560} & \textrm{---\,\textit{ibid}\,---} & \\
\expid{1773343565} & \textrm{---\,\textit{ibid}\,---} & \\
\expid{1773343568} & \textrm{---\,\textit{ibid}\,---} & \\
\expid{1773521107} & \textrm{---\,\textit{ibid}\,---} & \\
\expid{1773521116} & \textrm{---\,\textit{ibid}\,---} & \\
\expid{1773521128} & \textrm{---\,\textit{ibid}\,---} & \\
\expid{1773521135} & \textrm{---\,\textit{ibid}\,---} & \\
\expid{1773521150} & \textrm{---\,\textit{ibid}\,---} & \\
\expid{1773521156} & \textrm{---\,\textit{ibid}\,---} & \\
\expid{1773680909} & \textrm{---\,\textit{ibid}\,---} & \\

\expid{1780480777} & \textrm{---\,\textit{ibid}\,---} & \\
\expid{1780480782} & \textrm{---\,\textit{ibid}\,---} & \\
\expid{1780480787} & \textrm{---\,\textit{ibid}\,---} & \\
\expid{1780650581} & \textrm{---\,\textit{ibid}\,---} & \\
\expid{1780650584} & \textrm{---\,\textit{ibid}\,---} & \\
\expid{1780650586} & \textrm{---\,\textit{ibid}\,---} & \\
\expid{1780908507} & \textrm{---\,\textit{ibid}\,---} & \\
\expid{1780908509} & \textrm{---\,\textit{ibid}\,---} & \\
\expid{1780908511} & \textrm{---\,\textit{ibid}\,---} & \\
\expid{1781007723} & \textrm{---\,\textit{ibid}\,---} & \\
\expid{1781007725} & \textrm{---\,\textit{ibid}\,---} & \\
\expid{1781007727} & \textrm{---\,\textit{ibid}\,---} & \\
\expid{1781079773} & \textrm{---\,\textit{ibid}\,---} & \\
\expid{1781079775} & \textrm{---\,\textit{ibid}\,---} & \\
\expid{1781079777} & \textrm{---\,\textit{ibid}\,---} & \\
\expid{1781185330} & \textrm{---\,\textit{ibid}\,---} & \\
\expid{1781185333} & \textrm{---\,\textit{ibid}\,---} & \\
\expid{1781185335} & \textrm{---\,\textit{ibid}\,---} & \\
\expid{1781256852} & \textrm{---\,\textit{ibid}\,---} & \\
\expid{1781256854} & \textrm{---\,\textit{ibid}\,---} & \\
\expid{1781256856} & \textrm{---\,\textit{ibid}\,---} & \\
\expid{1781514760} & \textrm{---\,\textit{ibid}\,---} & \\
\expid{1781514764} & \textrm{---\,\textit{ibid}\,---} & \\
\expid{1781514767} & \textrm{---\,\textit{ibid}\,---} & \\
\expid{1781621786} & \textrm{---\,\textit{ibid}\,---} & \\
\expid{1781621789} & \textrm{---\,\textit{ibid}\,---} & \\
\expid{1781621793} & \textrm{---\,\textit{ibid}\,---} & \\
\expid{1781697075} & \textrm{---\,\textit{ibid}\,---} & \\
\expid{1781697078} & \textrm{---\,\textit{ibid}\,---} & \\
\expid{1781697081} & \textrm{---\,\textit{ibid}\,---} & \\
\expid{1781790629} & \textrm{---\,\textit{ibid}\,---} & \\
\expid{1781790631} & \textrm{---\,\textit{ibid}\,---} & \\
\expid{1781790633} & \textrm{---\,\textit{ibid}\,---} & \\
\expid{1781871082} & \textrm{---\,\textit{ibid}\,---} & \\
\expid{1781871084} & \textrm{---\,\textit{ibid}\,---} & \\
\expid{1781871086} & \textrm{---\,\textit{ibid}\,---} & \\
\expid{1782129149} & \textrm{---\,\textit{ibid}\,---} & \\
\expid{1782129152} & \textrm{---\,\textit{ibid}\,---} & \\
\expid{1782129154} & \textrm{---\,\textit{ibid}\,---} & \\
\expid{1782209424} & \textrm{---\,\textit{ibid}\,---} & \\
\expid{1782209429} & \textrm{---\,\textit{ibid}\,---} & \\
\expid{1782209437} & \textrm{---\,\textit{ibid}\,---} & \\
\expid{1782290038} & \textrm{---\,\textit{ibid}\,---} & \\
\expid{1782290044} & \textrm{---\,\textit{ibid}\,---} & \\
\expid{1782290047} & \textrm{---\,\textit{ibid}\,---} & \\
\expid{1782391919} & \textrm{---\,\textit{ibid}\,---} & \\
\expid{1782391921} & \textrm{---\,\textit{ibid}\,---} & \\
\expid{1782391924} & \textrm{---\,\textit{ibid}\,---} & \\
\expid{1782463369} & \textrm{---\,\textit{ibid}\,---} & \\
\expid{1782463371} & \textrm{---\,\textit{ibid}\,---} & \\
\expid{1782463373} & \textrm{---\,\textit{ibid}\,---} & \\
\expid{1782725684} & \textrm{---\,\textit{ibid}\,---} & \\
\expid{1782725686} & \textrm{---\,\textit{ibid}\,---} & \\
\expid{1782725688} & \textrm{---\,\textit{ibid}\,---} & \\
\expid{1782814250} & \textrm{---\,\textit{ibid}\,---} & \\
\expid{1782822554} & \textrm{---\,\textit{ibid}\,---} & \\
\expid{1782900715} & \textrm{---\,\textit{ibid}\,---} & \\
\expid{1782900776} & \textrm{---\,\textit{ibid}\,---} & \\
\expid{1782900741} & \textrm{---\,\textit{ibid}\,---} & \\
\expid{1782993286} & \textrm{---\,\textit{ibid}\,---} & \\
\expid{1782993287} & \textrm{---\,\textit{ibid}\,---} & \\
\expid{1782993289} & \textrm{---\,\textit{ibid}\,---} & \\
\expid{1783068960} & \textrm{---\,\textit{ibid}\,---} & \\
\expid{1783068962} & \textrm{---\,\textit{ibid}\,---} & \\
\expid{1783068964} & \textrm{---\,\textit{ibid}\,---} & \\
\expid{1783326897} & \textrm{---\,\textit{ibid}\,---} & \\
\expid{1783326899} & \textrm{---\,\textit{ibid}\,---} & \\
\expid{1783326902} & \textrm{---\,\textit{ibid}\,---} & \\
\expid{1783513609} & \textrm{---\,\textit{ibid}\,---} & \\
\expid{1783513611} & \textrm{---\,\textit{ibid}\,---} & \\
\expid{1783513612} & \textrm{---\,\textit{ibid}\,---} & \\

\end{longtable}
\end{scriptsize}

\subsection{Disk Failures}
Disk failure experiments simulate the loss of one or more NVMe data disks (\texttt{g1disk01}, \texttt{g1disk02}, \texttt{g2disk01}, \texttt{g2disk02}) on the storage servers.
Three subtypes are distinguished:
\emph{single} disk failures affect exactly one disk;
\emph{simultaneous} failures take multiple disks offline at the same instant;
\emph{asynchrnonous} failures spread across time and are further categorized as \emph{cascading} (failures start close together and mostly overlap), \emph{rolling} (sequential with partial overlap), or \emph{independent} (no temporal overlap).

In several simultaneous experiments one or more disks were not
successfully recovered after the planned fix and remained offline for the remainder of the experiment; these are noted as ``permanent loss.'' Subsequent experiments are run with a 10-minute staggered disk recovery strategy on the same server to avoid those events. Two of the experiments with permanent disk loss were rerun with this fix in effect. Experiments with permanent disk loss still produce perfectly labeled data; they simply contain much longer local anomalies.
\begin{scriptsize}
\begin{longtable}{l p{8cm} p{3cm}}
\caption{Disk failure experiments (38 runs).}
\label{tab:registry-disk} \\
\toprule
\textbf{ID} & \textbf{Details} & \textbf{Notes} \\
\midrule
\endfirsthead
\toprule
\textbf{ID} & \textbf{Details} & \textbf{Notes} \\
\midrule
\endhead
\bottomrule
\endfoot
\multicolumn{3}{l}{\textit{Single Disk Failures}} \\
\midrule
\expid{1773343796} & s1/g2disk01, onset 3h30m, dur 1h30m & \\
\expid{1773343802} & s3/g2disk02, onset 6h, dur 35m & \\
\expid{1773343833} & s3/g1disk02, onset 7h15m, dur 2h30m & \\
\expid{1773343804} & s1/g2disk01, onset 10h, dur 10m & \\
\expid{1773343837} & s1/g2disk02, onset 8h45m, dur 3h15m & \\
\expid{1773603342} & s2/g2disk01, onset 9h30m, dur 20m & \\
\expid{1773603348} & s1/g1disk01, onset 10h15m, dur 4h & \\
\expid{1773603351} & s1/g2disk01, onset 11h, dur 50m & \\
\expid{1773603363} & s2/g2disk01, onset 12h30m, dur 1h10m & \\
\expid{1773603383} & s3/g1disk02, onset 14h30m, dur 2h & \\
\midrule
\multicolumn{3}{l}{\textit{Simultaneous Multiple Disk Failures}} \\
\midrule
\expid{1773429303} & s1/g1disk01 + s2/g2disk01, onset 5h, dur 45m & \\
\expid{1773429328} & s3/g1disk01 + s3/g2disk02 (same server), onset 12h, dur 3h30m & g1disk01 permanent loss \\
\expid{1773429369} & s1/g2disk02 + s2/g1disk01 + s3/g2disk01, onset 10h, dur 1h30m & \\
\expid{1773429404} & s1/g1disk01 + s1/g1disk02 + s2/g2disk01 + s3/g1disk02, onset 7h30m, dur 30m & s1/g1disk01 permanent loss \\
\expid{1773680793} & s2/g1disk02 + s3/g2disk01, onset 8h, dur 2h & \\
\expid{1773680834} & s1/g2disk01 + s1/g2disk02, onset 14h, dur 1h & s1/g2disk02 permanent loss \\
\expid{1773680847} & s1/g1disk01 + s1/g2disk01 + s2/g2disk02, onset 3h, dur 2h & s1/g2disk01 permanent loss \\
\expid{1773680864} & s2/g1disk02 + s3/g1disk01 + s3/g2disk02, onset 15h, dur 45m & s3/g1disk01 permanent loss \\
\expid{1773680884} & s1/g1disk02 + s1/g2disk01 + s3/g1disk01 + s3/g2disk02, onset 6h, dur 1h15m & s1/g2disk01 + s3/g2disk02 permanent loss \\
\expid{1773680896} & s1/g2disk02 + s2/g1disk01 + s2/g2disk01 + s3/g1disk02, onset 11h, dur 3h & s2/g2disk01 permanent loss \\
\expid{1773767900} & s3/g1disk01 3h30m + s3/g2disk02 3h40m, onset 12h & rerun of \expref{1773429328} \\
\expid{1773768006} & s1/g1disk02 1h15m + s1/g2disk01 1h25m + s3/g1disk01 1h15m + s3/g2disk02 1h25m, onset 6h & rerun of \expref{1773680884} \\
\expid{1775270767} & s1/g1disk01 1h + s1/g1disk02 1h10m, onset 13h30m & \\
\expid{1775270771} & s1/g1disk02 1h50m + s1/g2disk02 2h + s2/g1disk01 1h50m, onset 3h30m & \\
\expid{1775270774} & s2/g2disk01 45m + s3/g1disk02 55m + s3/g2disk01 50m, onset 15h30m & \\
\expid{1775270776} & s1/g2disk01 35m + s1/g2disk02 45m + s2/g1disk02 35m + s3/g2disk01 40m, onset 7h & \\
\expid{1775270779} & s1/g1disk02 2h50m + s2/g2disk02 3h + s2/g1disk02 2h50m + s3/g1disk01 3h10m, onset 11h30m & \\
\midrule
\multicolumn{3}{l}{\textit{Asynchronous Multiple Disk Failures}} \\
\midrule
\expid{1773429559} & cascading: s1/g1disk01 @6h/1h; s2/g2disk01 @6h15m/50m & \\
\expid{1773429588} & cascading: s1/g1disk01 @4h/40m; s2/g1disk02 @4h8m/30m; s3/g2disk01 @4h20m/55m & s1/g1disk01 permanent loss \\
\expid{1773429647} & independent: s1/g2disk02 @9h/1h30m; s3/g1disk01 @12h/2h & \\
\expid{1773429670} & rolling: s1/g2disk01 @8h/3h; s2/g1disk01 @10h/2h; s2/g2disk02 @11h30m/1h15m; s3/g1disk02 @12h15m/1h & \\
\expid{1773767292} & cascading: s2/g2disk01 @13h/1h30m; s2/g1disk02 @13h10m/1h; s3/g2disk02 @13h25m/1h & \\
\expid{1773767367} & rolling: s1/g1disk02 @5h/2h; s3/g2disk01 @6h30m/1h30m & \\
\expid{1773767388} & rolling: s1/g2disk01 @2h/1h30m; s3/g1disk01 @3h/1h; s2/g2disk02 @3h30m/2h; s1/g1disk01 @4h15m/1h30m & \\
\expid{1773767435} & independent: s3/g2disk02 @3h/45m; s2/g1disk01 @15h/1h & \\
\expid{1773767461} & independent: s1/g2disk01 @2h/1h; s2/g2disk02 @7h/2h; s3/g1disk02 @14h/1h30m & \\
\expid{1773767785} & independent: s2/g1disk02 @1h30m/30m; s1/g2disk01 @5h/1h15m; s3/g1disk01 @10h/45m; s3/g2disk02 @15h/2h & \\
\expid{1775167457} & cascading: s3/g2disk01 @10h/1h15m; s1/g1disk02 @10h12m/45m & \\
\end{longtable}
\end{scriptsize}

\subsection{Server Failures}
Server failure experiments simulate the loss / degradation of an entire storage server.
Because the warp S3 load balancer distributes requests across all three servers in round-robin, losing one server reduces overall throughput proportionally and changes the traffic shape seen by the remaining servers. 
\emph{Single} instant failures drop all network traffic on the target server.
\emph{Asynchronous} instant failures apply the same mechanism to multiple servers at different times within a single experiment (two servers cannot be killed simultaneously).
\emph{Degradation} experiments inject progressive network impairment in two ways: 
\emph{linear} ramps delay and packet loss smoothly from zero to configured maximums over the anomaly duration, while
\emph{gradual} applies them in discrete stages at configured percentages of the anomaly duration, creating a staircase pattern.

Two experiments (\expref{1773946658} and \expref{1773946737}) experienced unplanned server crashes during execution, where the targeted server failed unpredictably due to stress caused by the anomaly and had to be manually reset after a much longer duration.
Both were subsequently re-run with shorter, controlled durations, and therefore very different telemetry profiles.

\begin{scriptsize}
\begin{longtable}{l p{8cm} p{3cm}}
\caption{Server failure experiments (48 runs).}
\label{tab:registry-server} \\
\toprule
\textbf{ID} & \textbf{Details} & \textbf{Notes} \\
\midrule
\endfirsthead
\toprule
\textbf{ID} & \textbf{Details} & \textbf{Notes} \\
\midrule
\endhead
\bottomrule
\endfoot
\multicolumn{3}{l}{\textit{Single Server Failures}} \\
\midrule
\expid{1773946658} & store 1, onset 6h, dur 9h45m & unplanned crash \\
\expid{1774026192} & store 1, onset 6h, dur 20m & rerun of \expref{1773946658} \\
\expid{1774026217} & store 2, onset 10h, dur 15m & \\
\expid{1774132105} & store 3, onset 3h, dur 5m & \\
\expid{1774132113} & store 2, onset 14h, dur 1m & \\
\expid{1774587691} & store 2, onset 7h, dur 25m & \\
\expid{1774587733} & store 1, onset 30m, dur 8m & \\
\expid{1775087951} & store 2, onset 5h, dur 22m & \\
\expid{1775087960} & store 3, onset 11h, dur 18m & \\
\expid{1775087966} & store 3, onset 16h, dur 3m & \\
\expid{1775087972} & store 1, onset 1h30m, dur 12m & \\
\midrule
\multicolumn{3}{l}{\textit{Asynchronous Instant Server Failures}} \\
\midrule
\expid{1773946737} & s1 @4h/30m; s2 @10h/5h27m & unplanned crash on s2 \\
\expid{1774026232} & s1 @4h/30m; s2 @10h/25m & rerun of \expref{1773946737} \\
\expid{1774026246} & s1 @3h/20m; s2 @8h/15m; s3 @14h/5m & \\
\expid{1774132120} & s1 @2h/10m; s3 @7h/15m & \\
\expid{1774132127} & s2 @6h/5m; s3 @13h/3m & \\
\expid{1774587774} & s2 @4h/12m; s2 @11h/20m & \\
\expid{1775087979} & s1 @5h/10m; s1 @14h30m/15m & \\
\expid{1775087985} & s3 @2h30m/15m; s1 @12h/10m & \\
\expid{1775087991} & s1 @8h/7m; s3 @15h/18m & \\
\expid{1775087997} & s2 @1h/5m; s3 @6h30m/20m; s1 @13h/12m & \\
\expid{1775167468} & s3 @9h30m/6m; s2 @16h/14m & \\
\midrule
\multicolumn{3}{l}{\textit{Server Degradation --- Linear}} \\
\midrule
\expid{1774026280} & store 1, onset 6h, dur 2h, max 500ms delay / 15\% loss & \\
\expid{1774132137} & store 2, onset 4h, dur 1h30m, max 200ms delay / 8\% loss & \\
\expid{1774132147} & store 1, onset 10h, dur 4h, max 750ms delay / 20\% loss & \\
\expid{1774291956} & store 3, onset 5h, dur 2h, max 15ms delay / 4\% loss & \\
\expid{1774291974} & store 2, onset 7h, dur 3h, max 40ms delay / 10\% loss & \\
\expid{1774479145} & store 1, onset 2h, dur 1h, max 2ms delay / 1\% loss & \\
\expid{1774479151} & store 2, onset 5h, dur 2h30m, max 4ms delay / 3\% loss & \\
\expid{1774479157} & store 3, onset 11h, dur 1h30m, max 5ms delay / 5\% loss & \\
\expid{1774479161} & store 1, onset 8h, dur 3h30m, max 3ms delay / 2\% loss & \\
\expid{1774586951} & store 3, onset 2h, dur 5h, max 1.5ms delay / 0.8\% loss & \\
\expid{1774586961} & store 1, onset 6h, dur 4h30m, max 0.8ms delay / 1\% loss & \\
\expid{1775270764} & store 2, onset 5h, dur 3h, max 0.8ms delay / 0\% loss & \\
\midrule
\multicolumn{3}{l}{\textit{Server Degradation --- Gradual}} \\
\midrule
\expid{1774026259} & store 1, onset 6h, dur 2h: 100ms/5\% @25\%, 300ms/10\% @50\%, 800ms/20\% @75\% & \\
\expid{1774026268} & store 2, onset 8h, dur 3h: 50ms/2\% @20\%, 150ms/12\% @40\%, 400ms/12\% @60\%, 1s/25\% @80\% & \\
\expid{1774132154} & store 3, onset 4h, dur 1h30m: 150ms/5\% @33\%, 500ms/15\% @66\% & \\
\expid{1774132163} & store 2, onset 10h, dur 4h: 20ms/1\% @10\%, 80ms/3\% @30\%, 200ms/8\% @50\%, 500ms/15\% @70\%, 1s/30\% @90\% & \\
\expid{1774291987} & store 1, onset 3h, dur 2h30m: 3ms/0.5\% @2\%, 5ms/1.5\% @25\%, 7ms/3\% @55\%, 10ms/5\% @85\% & \\
\expid{1774292005} & store 3, onset 9h, dur 3h: 5ms/1\% @3\%, 20ms/4\% @35\%, 35ms/7\% @70\% & \\
\expid{1774478710} & store 2, onset 3h, dur 2h: 1ms/0.5\% @3\%, 4ms/3\% @50\% & \\
\expid{1774478725} & store 3, onset 6h, dur 1h: 1ms/0.3\% @2\%, 3ms/1.5\% @35\%, 5ms/4\% @70\% & \\
\expid{1774478736} & store 1, onset 12h, dur 2h30m: 1ms/0.5\% @3\%, 2ms/1\% @25\%, 3ms/2.5\% @55\%, 5ms/5\% @80\% & \\
\expid{1774478743} & store 2, onset 5h, dur 3h: 0.2ms/0.2\% @2\%, 0.8ms/0.8\% @20\%, 1.2ms/1.5\% @40\%, 1.4ms/3\% @65\%, 1.6ms/4.5\% @90\% & \\
\expid{1774586933} & store 2, onset 10h, dur 4h: 0.3ms/0.1\% @3\%, 0.5ms/0.25\% @35\%, 0.7ms/0.4\% @70\% & \\
\expid{1774586939} & store 3, onset 7h, dur 6h: 0.1ms/0.05\% @2\%, 0.7ms/0.1\% @25\%, 0.5ms/0.15\% @55\%, 0.8ms/0.2\% @85\% & \\
\expid{1775270759} & store 1, onset 4h, dur 2h: 0.1ms/0.1\% @3\%, 0.3ms/0.3\% @40\%, 0.5ms/0.5\% @75\% & \\
\expid{1775270761} & store 3, onset 9h, dur 4h: 0.3ms/0.2\% @3\%, 0.7ms/0.5\% @30\%, 0.2ms/0.1\% @55\%, 0.5ms/0.4\% @80\% & \\
\end{longtable}
\end{scriptsize}

\subsection{Storage Node Failures}
Software/storage node (snode) experiments kill a configurable number of the 6 software storage nodes running on a single server.
The same three subtypes as disk failures are used:
\emph{single} failures kill snodes on one server;
\emph{simultaneous} failures kill snodes on multiple servers at the same time;
\emph{asynchronous} failures spread across servers over time, categorized as \emph{cascading} (overlapping onset), \emph{rolling} (sequential with partial overlap), or \emph{independent} (no temporal overlap).

16 of the 46 snode experiments were affected by a software bug caused by the anomaly, affecting the targeted server's telemetry agent and causing it to discard other server level metrics. 
Those experiments as well as their reruns are included, as the bug created a fully different telemetry profile for each affected time series.

\begin{scriptsize}
\begin{longtable}{l p{8cm} p{3cm}}
\caption{Storage node failure experiments (46 runs).}
\label{tab:registry-snode} \\
\toprule
\textbf{ID} & \textbf{Details} & \textbf{Notes} \\
\midrule
\endfirsthead
\toprule
\textbf{ID} & \textbf{Details} & \textbf{Notes} \\
\midrule
\endhead
\bottomrule
\endfoot
\multicolumn{3}{l}{\textit{Single-Server Software Node Failures}} \\
\midrule
\expid{1774656169} & 1 snode on s1, onset 6h, dur 1h & server metric loss \\
\expid{1774656174} & 1 snode on s2, onset 9h, dur 1h30m & server metric loss \\
\expid{1774656180} & 2 snodes on s1, onset 5h, dur 45m & server metric loss \\
\expid{1774656183} & 2 snodes on s3, onset 8h, dur 1h & server metric loss \\
\expid{1774764369} & 3 snodes on s2, onset 2h, dur 30m & server metric loss \\
\expid{1774764432} & 3 snodes on s3, onset 14h, dur 45m & server metric loss \\
\expid{1774844948} & 1 snode on s1, onset 6h, dur 1h & rerun of \expref{1774656169} \\
\expid{1774844958} & 1 snode on s2, onset 9h, dur 1h30m & rerun of \expref{1774656174} \\
\expid{1774844963} & 2 snodes on s1, onset 5h, dur 45m & rerun of \expref{1774656180} \\
\expid{1774844969} & 2 snodes on s3, onset 8h, dur 1h & rerun of \expref{1774656183} \\
\expid{1774936591} & 3 snodes on s2, onset 2h, dur 30m & rerun of \expref{1774764369} \\
\expid{1774936599} & 3 snodes on s3, onset 14h, dur 45m & rerun of \expref{1774764432} \\
\expid{1775021366} & 1 snode on s3, onset 16h, dur 20m & \\
\expid{1775021375} & 1 snode on s1, onset 3h, dur 15m & \\
\expid{1775167481} & 3 snodes on s1, onset 7h30m, dur 35m & \\
\expid{1775167484} & 2 snodes on s2, onset 12h, dur 50m & \\
\midrule
\multicolumn{3}{l}{\textit{Simultaneous Multi-Server Software Node Failures}} \\
\midrule
\expid{1774656573} & s1: 1 snode + s2: 1 snode, onset 13h, dur 1h & server metric loss \\
\expid{1774656577} & s1: 1 snode + s2: 2 snodes, onset 7h, dur 45m & server metric loss \\
\expid{1774764331} & s1: 1 snode 20m + s2: 1 snode 25m + s3: 1 snode 30m, onset 10h & server metric loss \\
\expid{1774764358} & s1: 2 snodes + s3: 2 snodes, onset 15h, dur 15m & server metric loss \\
\expid{1774844979} & s1: 1 snode + s2: 1 snode, onset 13h, dur 1h & rerun of \expref{1774656573} \\
\expid{1774844985} & s1: 1 snode + s2: 2 snodes, onset 7h, dur 45m & rerun of \expref{1774656577} \\
\expid{1774936575} & s1: 1 snode 20m + s2: 1 snode 25m + s3: 1 snode 30m, onset 10h & rerun of \expref{1774764331} \\
\expid{1774936581} & s1: 2 snodes + s3: 2 snodes, onset 15h, dur 15m & rerun of \expref{1774764358} \\
\expid{1775021334} & s2: 1 snode + s3: 1 snode, onset 3h30m, dur 20m & \\
\expid{1775021343} & s1: 1 snode + s3: 1 snode, onset 11h30m, dur 15m & \\
\expid{1775021351} & s1: 2 snodes + s2: 1 snode + s3: 2 snodes, onset 6h, dur 20m & \\
\expid{1775021360} & s2: 2 snodes 25m + s3: 1 snode 30m, onset 16h30m & \\
\expid{1775167475} & s2: 3 snodes 25m + s3: 1 snode 30m, onset 14h & \\
\expid{1775167478} & s1: 1 snode + s3: 2 snodes, onset 4h, dur 40m & \\
\midrule
\multicolumn{3}{l}{\textit{Non-Simultaneous Multi-Server Software Node Failures}} \\
\midrule
\expid{1774656607} & cascading: s1 2 snodes @3h30m/1h; s2 1 snode @4h/1h & server metric loss \\
\expid{1774656595} & independent: s1 1 snode @5h/1h; s3 2 snodes @11h/1h30m & server metric loss \\
\expid{1774764255} & cascading: s1 1 snode @3h/1h30m; s2 2 snodes @3h45m/1h; s3 1 snode @4h15m/35m & server metric loss \\
\expid{1774764269} & independent: s2 3 snodes @4h/45m; s3 2 snodes @13h/1h10m & server metric loss \\
\expid{1774764282} & rolling: s1 1 snode @9h/1h30m; s2 1 snode @11h/1h20m; s3 2 snodes @14h/1h & server metric loss \\
\expid{1774764299} & rolling: s1 2 snodes @2h/1h30m; s2 3 snodes @3h30m/1h; s3 1 snode @5h/2h; s1 1 snode @12h/1h & server metric loss \\
\expid{1774844991} & cascading: s1 2 snodes @3h30m/1h; s2 1 snode @4h/1h & rerun of \expref{1774656607} \\
\expid{1774844998} & independent: s1 1 snode @5h/1h; s3 2 snodes @11h/1h30m & rerun of \expref{1774656595} \\
\expid{1774936535} & cascading: s1 1 snode @3h/1h30m; s2 2 snodes @3h45m/1h; s3 1 snode @4h15m/35m & rerun of \expref{1774764255} \\
\expid{1774936551} & independent: s2 3 snodes @4h/45m; s3 2 snodes @13h/1h10m & rerun of \expref{1774764269} \\
\expid{1774936559} & rolling: s1 1 snode @9h/1h30m; s2 1 snode @11h/1h20m; s3 2 snodes @14h/1h & rerun of \expref{1774764282} \\
\expid{1774936567} & rolling: s1 2 snodes @2h/1h30m; s2 3 snodes @3h30m/1h; s3 1 snode @5h/2h; s1 1 snode @12h/1h & rerun of \expref{1774764299} \\
\expid{1775021320} & cascading: s3 1 snode @12h/1h; s1 2 snodes @12h45m/45m & \\
\expid{1775021329} & rolling: s2 1 snode @5h/20m; s3 1 snode @6h/15m; s1 1 snode @7h/25m & \\
\expid{1775167464} & independent: s1 2 snodes @2h30m/40m; s2 1 snode @15h/1h & \\
\expid{1775167472} & rolling: s3 1 snode @8h/1h; s1 2 snodes @8h30m/45m & \\
\end{longtable}
\vspace{1cm}
\end{scriptsize}

\section{Technical Details}
\label{app:tech_details}

\subsection{System Details}
\begin{figure}[ht]
    \centering
    \input{figures/topology}
    \caption{Experiment cluster topology. The metric hierarchy on the right shows all four telemetry levels and their dimension counts, totaling 171 features.}
    \label{fig:topology}
    \label{app:topology} \vspace{1cm}
\end{figure}
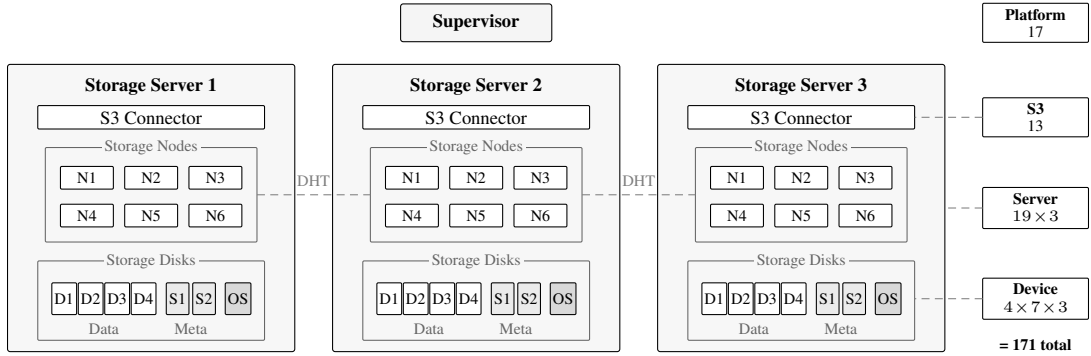

The cluster topology shared by all experimental deployments is illustrated in Figure~\ref{fig:topology}. Each cluster runs RING 9.5.0.5 on top of Rocky Linux 9.5 with 3 storage servers and one supervisor server. Each storage server is equipped with four 10\,GB NVMe data disks and two 5\,GB SSDs for metadata, in addition to one OS disk.

Client S3 traffic generated by warp on the supervisor enters each storage server through its S3 connector service. The connector parses S3 API calls, performs metadata lookups, and dispatches data operations to the appropriate storage nodes via the DHT. Warp distributes client requests to all three servers in round-robin, before further load balancing is done at the connector level.

The RING's data placement in each storage server is governed by an implementation of the Chord protocol~\cite{stoica2003chord}, a peer-to-peer distributed hash table in which object keys are hashed and mapped onto a logical ring of identifiers. Each of the 18 software storage nodes (6 per server) owns a contiguous arc of this keyspace and is responsible for storing all objects whose hashes fall within its range. Chord ensures that when a node joins or leaves the ring, only the keyspace segments adjacent to it are affected, as the protocol redistributes the affected key ranges to their new owners efficiently. Therefore, when storage nodes are killed, or servers are isolated, the remaining nodes absorb the orphaned keyspace segments and continue serving requests.

Objects are protected via replication or erasure coding depending on size, with replicas placed on distinct storage nodes across different servers for durability. Three categories of background tasks run continuously and are directly relevant to the anomaly experiments: \textit{rebuild} detects missing replicas or chunks (e.g., after a disk or node failure) and reconstructs them onto healthy disks; \textit{repair} scans for data integrity issues and fixes them from surviving copies; and \textit{rebalance} redistributes data when disk usage becomes uneven, typically after a recovery event. These tasks run at low intensity during nominal operation and ramp up after failure events, producing visible signatures in the per-server task-rate metrics (see Table~\ref{tab:metrics_server}).

\subsection{Implementation Details}
\label{app:implementation_details}

\paragraph{Baseline workload generation:}
S3 client traffic is generated by \texttt{warp}~v1.0.8\footnote{\url{https://github.com/minio/warp}}, an open-source S3 benchmarking tool. Warp runs on the supervisor server and distributes requests to all three storage servers via round-robin host selection (\texttt{-{}-host-select roundrobin}). The tool is invoked in stress mode (\texttt{-{}-stress}) to discard output for long-running experiments, and object sizes are randomized within the configured range (\texttt{-{}-obj.randsize}). An important consequence of the round-robin distribution is that, when a storage server becomes unreachable during a server failure experiment, approximately one-third of all S3 requests are dropped by the client, producing a visible and proportional throughput reduction across the cluster.

\paragraph{Disk failure injection:}
Disk failures are triggered by fully detaching the target NVMe disk from the operating system. The framework resolves the disk's mount point to its underlying block device and PCI bus address, then performs a lazy unmount and unbinds the NVMe driver via the kernel's \texttt{sysfs} interface. This removes the block device entirely, simulating physical hardware loss rather than merely blocking I/O.
Recovery rebinds the NVMe driver, waits for the device to reappear, marks the disk as out-of-service via the platform's disk management utility, wipes its partition signatures, and initiates a RING-level disk replacement. The RING then begins background rebuild tasks to reconstruct the lost replicas on healthy disks.
In a subset of simultaneous disk failure experiments, the replacement procedure failed silently and the target disk remained permanently offline for the rest of the experiment. These ``permanent disk loss'' cases are annotated in the experiment registry (Appendix~\ref{app:experiments}, Table~\ref{tab:registry-disk}) and represent a realistic failure mode. To mitigate this in other experiments, a staggered recovery strategy was adopted, where disks on the same server are fixed with slight time offsets to reduce contention during the replacement procedure.

\paragraph{Server failure injection:}
All server failure modes use Linux \texttt{tc} (traffic control) with the \texttt{netem} network emulation module to manipulate network traffic at the kernel level. SSH and ARP traffic is explicitly exempted from impairment throughout, preserving the experiment framework's remote management access to the target server.

In \textit{instant} mode, all non-SSH traffic is subjected to 100\% packet loss, effectively isolating the server from the rest of the cluster.
In \textit{linear} degradation mode, network delay and packet loss are gradually ramped from zero to their configured maximum values over the anomaly duration.
In \textit{gradual} (staged) degradation mode, delay and packet loss are applied in discrete steps at configured percentage milestones of the anomaly duration, producing a staircase pattern.
Recovery reverses all network impairments and restores the kernel's default queueing disciplines.
Two server failure experiments experienced genuine unplanned server crashes caused by the stress of the injected anomaly, requiring manual recovery after a much longer outage than planned. Both were subsequently re-run with shorter, controlled durations. The original and re-run experiments are both included in the dataset and annotated in the experiment registry (Appendix~\ref{app:experiments}, Table~\ref{tab:registry-server}).

\paragraph{Storage node failure injection:}
Storage node failures target individual software processes on a given server. The experiment framework randomly selects the configured number of targets (1--3 out of the 6 running per server) and terminates them with \texttt{SIGABRT} (signal~6), producing a core dump that mimics a real application crash.
Recovery restarts each killed node via the platform's process management wrapper, after which the RING initiates background rebuild and repair tasks to restore the node's data.
A key distinction from server failures is that the storage server itself remains reachable: its remaining storage nodes continue to handle requests normally, and only the DHT keyspace segments owned by the killed nodes become temporarily unavailable while the Chord protocol redistributes their workload.
16 early storage node experiments were affected by a telemetry agent bug: killing a storage node process caused an unhandled connection error in the server's local telemetry agent, which disrupted metric collection for all services on the affected server. The agent was subsequently patched, and the experiments were re-run. Both versions are included in the dataset, as they produce distinct telemetry signatures, and are annotated accordingly in the experiment registry (Appendix~\ref{app:experiments}, Table~\ref{tab:registry-snode}).

\paragraph{Experiment setup:}
Internally, experiments are defined declaratively as YAML configuration files specifying the workload parameters and a timeline of fault injection events, each with an absolute time offset from experiment start, a configured duration and subsequent anomaly-specific parameters. A distribution of anomaly parameter coverage across all experiments can be found in Table~\ref{tab:coverage}.
The orchestrator schedules each timeline event at its specified offset from experiment start. Anomalies are automatically reversed after their configured duration. Experiments end after the baseline workload has been running for exactly 18 hours.

\begin{table}[H]
\footnotesize
\centering
\caption{Anomaly parameters coverage in SHAD.  \textit{C/R/I} denotes \textit{cascading}, \textit{rolling}, and \textit{independent} subtypes, respectively.}
\label{tab:coverage}
\begin{tabularx}{\textwidth}{@{}lXXX@{}}
\toprule
Type (\textit{Subtype}) & Onset & Duration & Parameters \\
\midrule 
Nominal  & -- & 18\,h & normal behaviour \\
Single disk & 3.5--14.5\,h & 10--240 min & 3 servers, 4 disk types \\
Simult.\ disk  & 3--15.5\,h & 30--215 min & 2-4 disks, 1-3 servers  \\
Async.\ disk (\textit{C/R/I})  & 1.5--15\,h & 30--180 min& 2-4 events per exp. \\ \hline
Single server  & 0.5--16\,h & 1--25 min$*$ & 3 servers \\
Async. server & 1--16\,h & 3--30 min$*$ & 2-3 events, same-server \\
Degradation (\textit{Lin/Grad})  & 2--12\,h & 1--6 h & delay 0.5--1000 ms, loss 0-30 \\ \hline
Single Snode  & 2--16\,h & 15--90 min & 1--3 Snodes, 3 servers\% \\
Simult. Snode & 3.5--16.5\,h & 15--60 min & 2--5 Snodes, 2--3 servers \\
Async. Snode (\textit{C/R/I})  & 2--15\,h & 15--120 min & 2--4 events per exp. \\
\bottomrule
\end{tabularx}
    \begin{tablenotes}
        \centering
        \item $*$\, Both anomaly types exclude 2 unpredictable server failures with much longer duration than average; (585\,min) for Instant and (327\,min) for Async.
    \end{tablenotes}
\end{table}

\newpage
\section{Experimental Evaluation Details}
\label{app:experimental_evaluation}
\subsection{Anomaly Detection}

  \begin{table}[H]
    \scriptsize
    \centering
    \caption{List of anomaly detectors used in the experiments. \textbf{Legend} refers to the name used for detectors in Figure \ref{fig:Detection}. }
    \label{tab:detector_list}
    \begin{tabularx}{\textwidth}{c c c c p{5.5cm}}
      \toprule
      \textbf{Name} & \textbf{Legend} & \textbf{Supervision} & \textbf{Category} & \textbf{Description}   \\
      \toprule
      AutoEncoder \cite{sakurada2014anomaly} & AE & S & \textit{Prediction-based} & Projects data to the lower-dimensional latent space and then reconstruct it through the encoding-decoding phase, where anomalies are typically characterized by evident reconstruction deviations. \\
       OmniAnomaly \cite{Omnianomaly} & OA & S & \textit{Prediction-based} & Stochastic recurrent neural network, which captures the normal patterns of time series by learning their robust representations with key techniques such as stochastic variable connection and planar normalizing flow, reconstructs input data by the representations, and use the reconstruction probabilities to determine anomalies.\\
       USAD \cite{usad} & USAD & S & \textit{Prediction-based} & based on adversely trained autoencoders, and the anomaly score is the combination of discriminator and reconstruction loss. \\
       CNN \cite{munir2018deepant} & CNN & S & \textit{Prediction-based} & Employs Convolutional Neural Network (CNN) to predict the next time stamp on the defined horizon and then compare the difference with the original value. \\
       LSTM \cite{lstm} & LSTM & S & \textit{Prediction-based} & Utilizes Long Short-Term Memory (LSTM) networks to model the relationship between current and preceding time series data, detecting anomalies through discrepancies between predicted and actual values. \\
       OCSVM \cite{scholkopf1999support} & SVM & S & \textit{Density-based} & Fits the dataset to find the normal data's boundary by maximizing the margin between the origin and the normal samples. \\
       TimesNet \cite{wu2022timesnet} & TN & S & \textit{Prediction-based} & A general time series analysis model with applications in forecasting, classification, and anomaly detection. It features TimesBlock, which can discover the multi-periodicity adaptively and extract the complex temporal variations from transformed 2D tensors by a parameter-efficient inception block. \\
       FITS \cite{xu2023fits} & FITS & S & \textit{Prediction-based} & Lightweight model that operates on the principle that time series can be manipulated through interpolation in the complex frequency domain. \\
       MCD \cite{mcd} & MCD & S & \textit{Density-based} & Based on minimum covariance determinant, which seeks to find a subset of all the sequences to estimate the mean and covariance matrix of the subset with minimal determinant. Subsequently, Mahalanobis distance is utilized to calculate the distance from sub-sequences to the mean, which is regarded as the anomaly score. \\
       Anomaly Transformer \cite{xu2021anomaly} & AT & S & \textit{Prediction-based} & Utilizes the `Anomaly-Attention' mechanism to compute the association discrepancy. \\
        \midrule
        KMeansAD \cite{yairi2001fault} & KMAD & U & \textit{Distance-based} & calculates the anomaly scores for each sub-sequence by measuring the distance to the centroid of its assigned cluster, as determined by the k-means algorithm. \\
        PCA \cite{aggarwal2015outlier} & PCA & U & \textit{Density-based} & Projects data to a lower-dimensional hyperplane, with significant deviation from this plane indicating potential outliers. \\
        CBLOF \cite{CBLOF} & CBLOF & U & \textit{Distance-based} & Cluster-based LOF, which calculates the anomaly score by first assigning samples to clusters, and then using the distance among clusters as anomaly scores. \\
        KNN \cite{knn} & KNN & U & \textit{Distance-based} & Produces the anomaly score of the input instance as the distance to its k-th nearest neighbor. \\
        HBOS \cite{hbos} & HBOS & U & \textit{Density-based} & Constructs a histogram for the data and uses the inverse of the height of the bin as the anomaly score of the data point. \\
        LOF \cite{LOF} & LOF & U & \textit{Distance-based} & Calculates the anomaly score by comparing local density with that of its neighbors. \\
        IsolationForest \cite{liu2008isolation} & IF & U & \textit{Density-based} &  Isolates anomalies using random partitioning of the data space. The core assumption is that anomalies are easier to isolate, requiring fewer random splits. Therefore, instances that require fewer partitions are more likely to be anomalous.  \\
        Robust PCA \cite{paffenroth2018robust} & RPCA & U & \textit{Density-based} & Built upon PCA, identifies anomalies by recovering the principal matrix. \\

      \bottomrule
    \end{tabularx}
    \begin{tablenotes}
        \scriptsize
        \centering
        \item \textbf{S}=Semi-Supervised $\;$ \textbf{U}=Unsupervised
    \end{tablenotes}
  \end{table}
  
Table \ref{tab:detector_list} shows the list of anomaly detectors used for the experimental evaluation in Section \ref{sec:experiments_tsad}. This pool of detectors is comprehensive and heterogeneous, and can be categorized according to the level of supervision (\textit{semi-supervised} or \textit{unsupervised}) or according to the underlying methodology (\textit{prediction-based}, \textit{density-based}, \textit{distance-based}).

\subsection{Prompt Template example}
\label{app:prompt}
\begin{figure}[t]
\centering
\includegraphics[width=\textwidth]{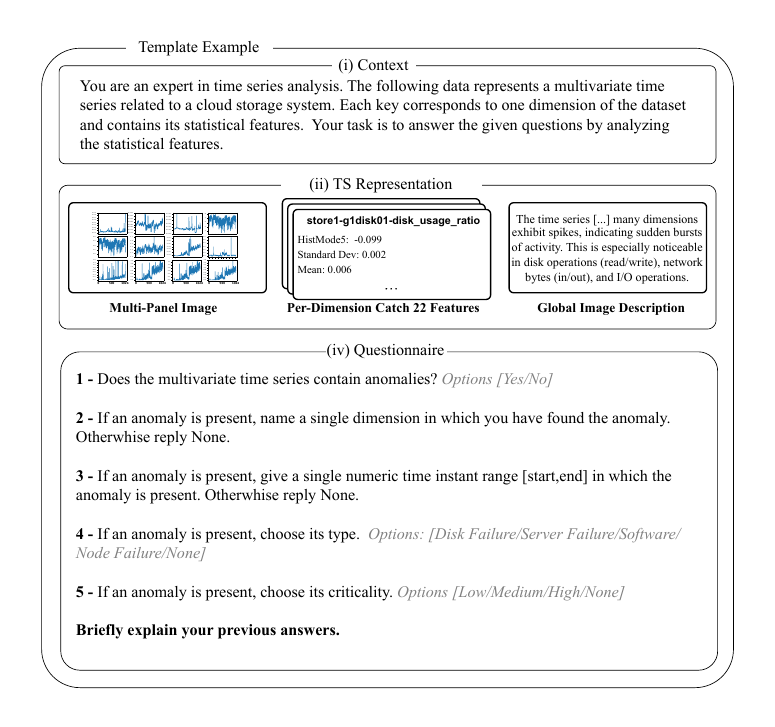}
\caption{Prompt template overview}
\label{fig:Prompt}
\end{figure}

Figure \ref{fig:Prompt} summarizes the prompt template used for the \textit{interpretability} study. The prompt is composed of three main elements: a (i) short description of the context and the task (the Figure shows the one used for Catch22), (ii) the provided representation, and (iii) the questionnaire. In addition, the LLM is asked to motivate its answers.  
The full prompts, available on our repository, also includes some formatting instructions to ensure the proper parsing of the LLM's answers. 

\subsection{Computing resources}

\subsubsection{Anomaly Detection}

\begin{figure}[t]
\centering
\includegraphics[width=0.8\textwidth]{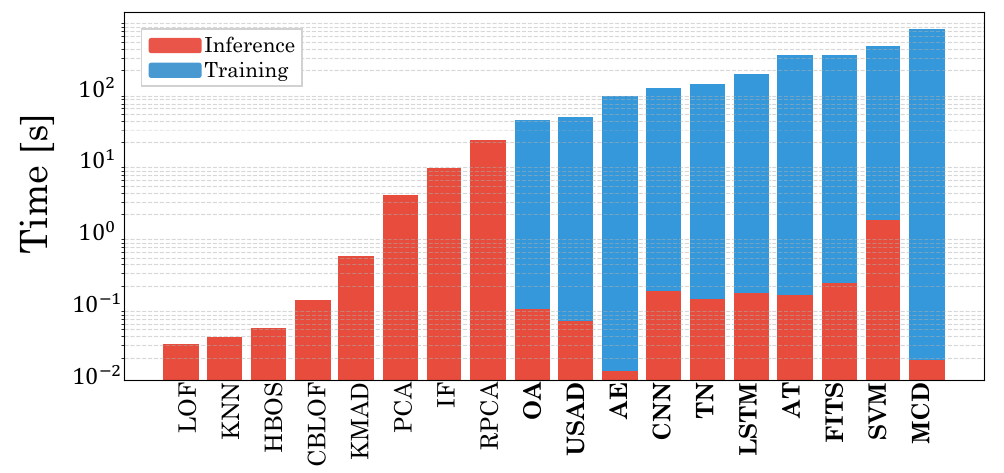}
\caption{Average Inference and Training Time on a Time Series for Each Method}
\label{fig:TimeAD}
\end{figure}

Figure~\ref{fig:TimeAD} reports the average time taken to obtain the anomaly score on a time series of the \textbf{SHAD} dataset. For Semi Supervised methods (in bold in the Figure) we report both the training time and the inference time. One major observation is that, understandingly, most of the Unsupervised methods are significantly faster than Semi-Supervised ones, as they have no training needed. Hence, with regard to the performances described in Section~\ref{sec:experiments_tsad}, KMAD appears as the best trade-off between performance and computational efficiency.

Non-deterministic methods were run five times, and the results reported are the average VUS-PR of those runs. The slowest method, FITS, takes a day ton run on the whole dataset. 

We conduct our experiments on a high-performance compute cluster with a heterogeneous configuration. The compute nodes feature a variety of architectures, including AMD EPYC (Zen 2/3) and Intel Xeon (Broadwell/Cascade Lake) processors, ranging from 16 to 64 physical cores. For GPU-accelerated methods, nodes are equipped with NVIDIA RTX 6000/8000 (24GB/48GB) or NVIDIA A100 (80GB) GPUs. All nodes operate on CentOS 7.9 (64-bit).

\subsubsection{Interpretability}
For the interpretability experiments, we rely on a large language model accessed via API. As inference is performed remotely, execution times are primarily influenced by network latency and the service response time, rather than local computational resources. In general, the average execution time of the full interpretability pipeline on a single dataset is 90 seconds for Catch22, 8 seconds for Images, and 14 seconds for Image Descriptions.  These timings may vary depending on external factors such as API load and network conditions, but provide a reasonable estimate of the computational overhead associated with LLM-based interpretation in our setting.

\subsection{Complete Reults tables}
Table \ref{tab:anomaly_types_explainability} reports a more stratified visualization of the explainability results of the top performing detectors (CNN, KMeansAd, and OmniAnomaly) across the different anomaly types. The results are also compared with the ones obtained with a random baseline. Overall, the results confirm the observation made in the main paper, i.e., that the number of affected dimensions can influence the intrinsic difficulty of the localization task. This is evident for highly localized anomalies (e.g., Single Disk). It is also noticeable that the NDCG values of the detectors are close to a random prediction, highlighting the complexity of the proposed dataset. 
 
Table \ref{tab:model_ranking} reports the average performance of each model across all representations, ranking them for each task. As it is possible to see, all models showed similar behaviours. Nevertheless, it is possible to notice that Claude was the best overall model, outperforming the others in 3 tasks (\textit{anomaly criticality}, \textit{temporal localization}, and \textit{anomaly type}) and resulting the second-best one for \textit{anomaly presence}.

\begin{table}[htbp]
\centering
\caption{Explainability results of top performing detectors and a random baseline for each anomaly type. The table also reports the average number of dimension affected by each anomaly type.}
\label{tab:anomaly_types_explainability}
\begin{tabular}{lrrrrrr}
\toprule
\textbf{Anomaly Type} & \textbf{Avg. Anomalous Dim.} & \textbf{CNN} & \textbf{KmeansAD} & \textbf{OmniAnomaly} & \textbf{Rand.} \\
\midrule
Single Disk          & 4      & 0.00 & 0.00 & 0.00 & 0.02 \\
Single Server        & 47     & 0.27 & 0.16 & 0.17 & 0.26 \\
Single Node          & 47     & 0.24 & 0.18 & 0.29 & 0.26 \\
Simultaneous Disk    & 12     & 0.05 & 0.11 & 0.00 & 0.08 \\
Simultaneous Node    & 100.7  & 0.57 & 0.46 & 0.60 & 0.57 \\
Asynchronous Disk    & 11.27  & 0.02 & 0.07 & 0.00 & 0.05 \\
Asynchronous Server  & 94     & 0.27 & 0.23 & 0.18 & 0.27 \\
Asynchronous Node    & 114.56 & 0.38 & 0.35 & 0.42 & 0.40 \\
Server Degradation   & 47     & 0.27 & 0.25 & 0.24 & 0.26 \\
\bottomrule
\end{tabular}
\vspace{1cm}
\end{table}

\begin{table}[t]
\centering
\caption{Model ranking by task. The performance is averaged across the three different representations (i.e., Catch 22, Images, and Descriptions).}
\begin{tabular}{llr}
\toprule
\textbf{Task} & \textbf{Model} & \textbf{Performance} \\
\midrule
Anomaly Criticality               & Claude   & 0.196400 \\
Anomaly Criticality               & Gemini   & 0.158063 \\
Anomaly Criticality               & Llama    & 0.149613 \\
Anomaly Criticality               & Mistral  & 0.147306 \\
Anomaly Criticality               & Grok     & 0.134204 \\
\midrule
Dimension Localization  & Grok     & 0.047685 \\
Dimension Localization  & Mistral  & 0.018056 \\
Dimension Localization  & Llama    & 0.008796 \\
Dimension Localization  & Gemini   & 0.006481 \\
Dimension Localization  & Claude   & 0.005556 \\
\midrule
Anomaly Presence                  & Gemini   & 0.493165 \\
Anomaly Presence                  & Claude   & 0.478134 \\
Anomaly Presence                  & Mistral  & 0.478134 \\
Anomaly Presence                  & Llama    & 0.476784 \\
Anomaly Presence                  & Grok     & 0.407620 \\
\midrule
Temporal Localization     & Claude   & 0.119444 \\
Temporal Localization     & Grok     & 0.047685 \\
Temporal Localization     & Gemini   & 0.047222 \\
Temporal Localization     & Mistral  & 0.031481 \\
Temporal Localization     & Llama    & 0.030556 \\
\midrule
Anomaly Type                      & Claude   & 0.235674 \\
Anomaly Type                      & Gemini   & 0.219983 \\
Anomaly Type                      & Mistral  & 0.183540 \\
Anomaly Type                      & Llama    & 0.176381 \\
Anomaly Type                      & Grok     & 0.101414 \\
\bottomrule
\end{tabular}
\label{tab:model_ranking}
\end{table}

Table \ref{tab:mean_gain} reports the average gain provided by each level of supervision, averaged across all models and representations. As it is possible to see, the best improvements concerns \textit{dimension localization} and \textit{temporal localization}.

\begin{table}[t]
\centering
\caption{Mean gain by supervision level and task.}
\begin{tabular}{llr}
\toprule
\textbf{Supervision} & \textbf{Task} & \textbf{Mean Gain} \\
\midrule
Level 1 & Anomaly Criticality               & \textcolor{darkgreen}{0.0128} \\
Level 2 & Anomaly Criticality               & \textcolor{red}{-0.0078} \\
Level 1 & Dimension Localization  & \textcolor{darkgreen}{\textbf{0.0719}} \\
Level 2 & Dimension Localization  & N.A. (trivial) \\
Level 1 & Anomaly Presence                  & N.A. (trivial) \\
Level 2 & Anomaly Presence                  & N.A. (trivial) \\
Level 1 & Temporal Localization     & \textcolor{darkgreen}{\textbf{0.0793}} \\
Level 2 & Temporal Localization     & \textcolor{darkgreen}{\textbf{0.1446}} \\
Level 1 & Anomaly Type                      & \textcolor{darkgreen}{0.0218} \\
Level 2 & Anomaly Type                      & \textcolor{darkgreen}{0.0477} \\
\bottomrule
\end{tabular}
\label{tab:mean_gain}
\end{table}

\section{Limitations}
\label{app:Limitations}
One of the limitations of SHAD is that each time series considers one type of anomalies, thereby not providing series with mixed failures (e.g. a server failure followed by a disk failure). Moreover, SHAD currently focuses on a single industrial distributed storage system. This design choice was made to prioritize annotation quality and reliability through close collaboration with domain experts, at the cost of reduced system heterogeneity. Extending the benchmark to additional storage systems and application domains is an important direction for future work, although it requires access to real infrastructures and substantial annotation effort. For the explainability experiment, we considered a traditional explainability method to understand whether simple dimension score attribution could be used to correctly identify anomalous dimensions. The evaluation of more refined strategies, which proved to be necessary in light of the results, is therefore left as future development. In addition, for evaluating interpretability, we provide a simple baseline consisting in a single-call solution. While more refined strategies (e.g., per-dimension analysis, advanced prompting, or multi-stage reasoning) or tailored approaches (e.g., ChatTS) can be used, the purpose of our experiments is to establish a simple baseline from which future work on LLM-based anomaly interpretation can build.

\section{Broader Impacts}
\label{app:impacts}
SHAD provides a comprehensive benchmark for time series anomaly detection, explanation, and interpretation on real-world, high-dimensional telemetry data from cloud storage systems. By enabling rigorous and reproducible evaluation of detection, explainability, and interpretability methods, SHAD can contribute to the development of more reliable and transparent monitoring systems. This may improve the robustness and availability of critical digital infrastructures (e.g., cloud services), leading to reduced downtime, better resource utilization, and improved user experience. Furthermore, the inclusion of per-dimension anomaly labels and rich textual annotations facilitates research on interpretable machine learning, helping practitioners better understand system failures and supporting faster debugging and root-cause analysis. This can reduce operational costs and lower the barrier to adopting advanced anomaly detection techniques in industry.

At the same time, models developed and evaluated exclusively using SHAD may exhibit limited generalization if applied to domains with substantially different characteristics, and anomaly detection systems may still produce false positives or missed detections in practice. Additionally, advances in system observability could, in certain contexts, be leveraged to infer sensitive operational patterns or to design strategies that evade detection. These considerations depend on specific deployment settings and highlight the importance of careful evaluation, domain-specific validation, and responsible use.

%% file: figures/topology.tex
\begin{tikzpicture}[
    every node/.style={font=\footnotesize, outer sep=0pt},
    server/.style={draw, rounded corners=1pt, minimum width=3.8cm, minimum height=3.8cm,
                   anchor=south west, fill=black!3, inner sep=2pt, outer sep=0pt},
    comp/.style={draw, rounded corners=0.5pt, minimum height=0.3cm, font=\scriptsize,
                 fill=white, inner sep=2pt, outer sep=0pt},
    sn/.style={draw, rounded corners=0.5pt, minimum width=0.7cm, minimum height=0.3cm,
               font=\tiny, fill=white, inner sep=1pt, outer sep=0pt},
    dk/.style={draw, rounded corners=0.5pt, minimum width=0.3cm, minimum height=0.42cm,
               font=\tiny, fill=white, inner sep=1pt, outer sep=0pt},
    dkm/.style={dk, fill=black!8},
    dko/.style={dk, fill=black!15},
    lbl/.style={font=\tiny, text=black!60, outer sep=0pt},
    metric/.style={draw, rounded corners=0.5pt, minimum width=1.4cm, minimum height=0.5cm,
                   font=\tiny, fill=white, inner sep=2pt, outer sep=0pt, anchor=west, align=center},
  ]

  \node[server] (s0) at (0,0) {};
  \node[server] (s1) at ([xshift=5mm]s0.south east) {};
  \node[server] (s2) at ([xshift=5mm]s1.south east) {};

  \node[draw, rounded corners=1pt, minimum width=2.0cm, minimum height=0.5cm,
        fill=black!4, inner sep=2pt, outer sep=0pt, anchor=south] (sup)
        at ([yshift=3mm]s1.north) {};
  \node[font=\scriptsize\bfseries] at (sup.center) {Supervisor};

  \foreach \s/\name in {s0/Storage Server 1, s1/Storage Server 2, s2/Storage Server 3} {
    \node[font=\scriptsize\bfseries] at (\s.north) [below=2pt] {\name};

    \node[comp, minimum width=3.0cm] at ([yshift=-7mm]\s.north) {S3 Connector};

    \draw[rounded corners=0.5pt, thin, black!60]
      ([xshift=5mm, yshift=14.5mm]\s.south west) rectangle ([xshift=33mm, yshift=27mm]\s.south west);
    \node[font=\tiny, text=black!60, fill=black!3, inner sep=1pt]
      at ([xshift=19mm, yshift=27mm]\s.south west) {Storage Nodes};

    \foreach \r in {0,1} {
      \foreach \c in {0,1,2} {
        \pgfmathtruncatemacro{\n}{\r*3+\c+1}
        \pgfmathsetmacro{\sxoff}{10.5 + \c*8.5}
        \pgfmathsetmacro{\syoff}{23.0 - \r*5.0}
        \node[sn] at ([xshift=\sxoff mm, yshift=\syoff mm]\s.south west) {N\n};
      }
    }

    \draw[rounded corners=0.5pt, thin, black!60]
      ([xshift=4mm, yshift=1.5mm]\s.south west) rectangle ([xshift=34mm, yshift=12mm]\s.south west);
    \node[font=\tiny, text=black!60, fill=black!3, inner sep=1pt]
      at ([xshift=19mm, yshift=12mm]\s.south west) {Storage Disks};

    \foreach \d/\l in {0/D1,1/D2,2/D3,3/D4} {
      \pgfmathsetmacro{\dxoff}{7.5 + \d*3.5}
      \node[dk] at ([xshift=\dxoff mm, yshift=7mm]\s.south west) {\l};
    }
    \foreach \d/\l in {0/S1,1/S2} {
      \pgfmathsetmacro{\dxoff}{22.5 + \d*3.5}
      \node[dkm] at ([xshift=\dxoff mm, yshift=7mm]\s.south west) {\l};
    }
    \node[dko] at ([xshift=30.5mm, yshift=7mm]\s.south west) {OS};

    \node[lbl] at ([xshift=12.75mm, yshift=3mm]\s.south west) {Data};
    \node[lbl] at ([xshift=24.25mm, yshift=3mm]\s.south west) {Meta};
  }

  \foreach \a/\b in {s0/s1, s1/s2} {
    \draw[thin, densely dashed, black!50]
      ([xshift=33mm, yshift=20.75mm]\a.south west) -- ([xshift=5mm, yshift=20.75mm]\b.south west)
      node[midway, above, font=\tiny, text=black!50] {DHT};
  }

  \node[metric] (m-plat) at ([xshift=5mm]s2.east |- sup.center)
        {\textbf{Platform}\\[-0.5pt]17};
  \node[metric] (m-s3) at ([xshift=5mm, yshift=31mm]s2.south east)
        {\textbf{S3}\\[-0.5pt]13};
  \node[metric] (m-srv) at ([xshift=5mm, yshift=19mm]s2.south east)
        {\textbf{Server}\\[-0.5pt]$19\!\times\!3$};
  \node[metric] (m-dev) at ([xshift=5mm, yshift=7mm]s2.south east)
        {\textbf{Device}\\[-0.5pt]$4\!\times\!7\!\times\!3$};

  \draw[thin, densely dashed, black!50]
    (m-s3.west) -- ([xshift=-4mm, yshift=31mm]s2.south east);
  \draw[thin, densely dashed, black!50]
    (m-srv.west) -- (s2.east);
  \draw[thin, densely dashed, black!50]
    (m-dev.west) -- ([xshift=-4mm, yshift=7mm]s2.south east);

  \node[font=\tiny\bfseries, anchor=north]
    at ([yshift=-1.5mm]m-dev.south) {= 171 total};

\end{tikzpicture}